\documentclass[preprint,12pt]{elsarticle}

\usepackage{amssymb}
\usepackage{amsmath}
\usepackage{graphicx}
\usepackage{algorithm}
\usepackage{multirow}
\usepackage{array}
\usepackage{amsmath}
\usepackage{mathtools}

\usepackage[noend]{algpseudocode}

\usepackage{bbm}
\usepackage{wrapfig,lipsum,booktabs}
\usepackage{amsfonts}
\newcolumntype{P}[1]{>{\centering\arraybackslash}p{#1}}
\newcolumntype{L}{>{\centering\arraybackslash}m{3cm}}
\usepackage{newfloat}
\usepackage{listings}
\usepackage{bm}
\usepackage{microtype}
\usepackage{subfigure}
\usepackage{booktabs} 
\usepackage{caption}
\usepackage{url}
\usepackage{lscape} 
\usepackage{longtable}
\usepackage{adjustbox}
\usepackage[dvipsnames]{xcolor}

\newcommand{\cN}{\mathcal{N}}

\newtheorem{thm}{Theorem}

\newtheorem{defn}[thm]{Definition}

\definecolor{rff}{RGB}{162, 21, 160}
\definecolor{trainable}{RGB}{33, 144, 33}

\journal{Nuerocomputing}

\begin{document}

\begin{frontmatter}



\title{An Explainable Gaussian Process Auto-encoder for Tabular Data} 



\author{Wei Zhang} 

\affiliation{organization={Electrical Engineering, Columbia University},
            addressline={116th and Broadway}, 
            city={New York},
            postcode={10025}, 
            state={NY},
            country={U.S.A.}}

\author{Brian Barr} 

\affiliation{organization={Capital One},
            addressline={1680 Capital One Dr}, 
            city={Mc Lean},
            postcode={22102}, 
            state={VA},
            country={U.S.A.}}

\author{John Paisley} 

\affiliation{organization={Electrical Engineering, Columbia University},
            addressline={116th and Broadway}, 
            city={New York},
            postcode={10025}, 
            state={NY},
            country={U.S.A.}}

\begin{abstract}
Explainable machine learning has attracted much interest in the community where the stakes are high. Counterfactual explanations methods have become an important tool in explaining a black-box model. The recent advances have leveraged the power of generative models such as an autoencoder. In this paper, we propose a novel method using a Gaussian process to construct the auto-encoder architecture for generating counterfactual samples. The resulting model requires fewer learnable parameters and thus is less prone to overfitting. We also introduce a novel density estimator that allows for searching for in-distribution samples. Furthermore, we introduce an algorithm for selecting the optimal regularization rate on density estimator while searching for counterfactuals. We experiment with our method in several large-scale tabular datasets and compare with other auto-encoder-based methods. The results show that our method is capable of generating diversified and in-distribution counterfactual samples. 
\end{abstract}

\begin{highlights}
\item We propose a novel framework for a counterfactual explainer using Gaussian process under an Auto-encoder architecture. We leverage Random Fourier feature approximation to reduce the computational cost. 
\item We also propose a novel density estimator in the latent space. The learned density estimator is deployed during the searching phase for counterfactual samples. The learned density estimator would direct the generated samples to be in-distribution of the target class. 
\item We experiment on five large-scale tabular datasets and evaluate our method with recent baselines under several widely-adopted counterfactual metrics. 
\end{highlights}


\end{frontmatter}



\section{Introduction}

Deep learning has been revolutionizing many fields such as computer vision, natural language processing and robotics. The recent development in natural language processing has shown that deep neural networks can achieve phenomenal performance when trained with a tremendous amount of data. Researchers usually recognize deep neural networks as a function approximation that maps an input to a meaningful representation, resulting in a better decision boundary for a task such as classification. Undoubtedly, efficient learning ability has advanced many fields. Though effective, the non-linearity of the deep models makes optimization non-convex and very difficult to explain, raising concerns about the fairness of the decision. This prevents its adoption in many high-stack fields such as healthcare and finance. This is called the black-box property of deep neural networks.

To mitigate this, many works take advantage of Generalized Additive Models(GAMs) to construct a white-box model that is transparent and self-explainable. The GAMs allow each input feature to have its own shape function and also extend to cross-pair features. The GAMs can be naturally explained by plotting the shape functions, showing how each feature would affect the outcome of the prediction. \citet{agarwal2021neural} first construct each shape function as a small neural network and called the resulting model Neural Additive Models (NAMs). \citet{radenovic2022neural} introduces basis functions to NAMs to reduce the trainable parameters. \citet{zhang2024gaussian} further reduces the number of parameters by deploying Gaussian processes with random Fourier features and makes the objective function convex.

On the other hand, researchers have proposed many post-hoc methods to explain a black-box model'decision through weight importance. \citet{ribeiro2016should} proposes Local Interpretable Model-Agnostic Explanations (LIME) to construct a linear model around the data points to approximate the black-box model's projection and use these learned linear weights to explain the outcome of the black-box model. Although the explanations are promising, such a method often suffers from the consistency issue for the provided explanation. \citet{sundararajan2017axiomatic, shrikumar2017learning} also propose the post-hoc method by comparing the query data point and a reference data point based on domain knowledge. The explanation is then obtained by calculating the gradients or activation on query data and reference data, showing which subset of input features are more important. Instead of a local explanation, \citet{ibrahim2019global} cluster the data points pointing to the decision boundary, from which global attributions are established. 

Recently, counterfactual explanation, first mentioned in \citet{pearl2009causality}, is a post-hoc method and is considered the highest level of the hierarchy of causal models. It usually requires imagining a reality that is opposite to the observed facts. Counterfactual explanations(CEs) intend to address the fundamental question: 'What modifications to the input are necessary to alter its outcome?' CEs aim to explain a classifier $f:\mathbb{R}^d\rightarrow \lbrace 0,1\rbrace$ by generating a counterfactual sample $\hat{x}$ such that the predicted label is flipped with minimal changes to the input ad defined by a metric $d(\cdot,\cdot)$. This can be characteristically formulated as
\begin{equation}\label{eqn:char_eqn}
    \arg\min_{\widehat{x}} d(x, \hat{x}) \quad \text{subject to} \quad f(\widehat{x}) = y_{\mathrm{target}}.
\end{equation}
Unlike other post-hoc methods, counterfactual inference uniquely offers the capability to determine these necessary changes, making it an effective tool in model interpretation and fairness evaluations. Consequently, counterfactual explanations not only generate a feature importance map but also provide tailored recommendations of adjustments to alter the outcome towards the target class. Counterfactual explanations can also be used to detect an unfair system if non-actionable features are proposed to change. 

A natural solution for generating counterfactual samples stems from using a generative model such as Variational autoencoders (VAEs) \cite{kingma2013auto}. Indeed, VAEs have been successfully applied to various tasks, including anomaly detection \cite{an2015variational}, temporal tracking \cite{zhang2018deep}, image captioning \cite{pu2016variational}, and the generation of arithmetic expressions \cite{kusner2017grammar}. Benefiting from its generative capabilities and scalability, the VAE has emerged as a popular choice for generating counterfactuals, prompting several extensions of its foundational framework \cite{pawelczyk2020learning,joshi2019towards,downs2020cruds,antoran2020getting}. Further details on these methods are discussed in Section 2. Although our framework is also built upon the generic auto-encoder architecture, there are three key distinctions between their frameworks and ours. First, our model unifies the classifier and AEs by supervising the latent space of the auto-encoder. The resulting encoder with a linear mapping is a deep classifier. It has been shown that adding a decoder as an unsupervised auxiliary task
will help learn a more distinguishable latent space\cite{le2018supervised}. Secondly, our model does not depend on the decoder for generating a counterfactual sample. The decoder, usually a deep neural network, introduces more uncertainty while generating samples and can rarely take immutable features into consideration. Instead, we directly back-propagate the gradients into the feature space, which enables us to mask immutable features. Last but not least, we leverage Gaussian process with Random Fourier features approximation for encoder and decoder and require fewer trainable parameters, making the entire framework less prone to overfitting. 

The generative power of a VAE heavily relies on the modeling capacity of its encoder and decoder, which are usually deep neural nets. In our proposed framework, we propose to adopt Gaussian processes (GPs) because they are widely recognized as a powerful Bayesian non-parametric approach for nonlinear modeling. Interestingly, GPs have gained attention as an approximation of infinitely wide deep neural networks \cite{lee2017deep}, with key findings indicating that GP predictions can surpass the performance of finite-width neural networks. However, the computational cost remains a significant challenge, as inverting the kernel typically requires $\mathcal{O}(N^3)$ time complexity. This bottleneck limits the applicability of GPs to large-scale datasets. To address this issue, \citet{titsias2009variational} introduced sparse GPs, utilizing KL divergence to select inducing points for kernel matrix approximation. The number of inducing points $m$ is much less than the number of data points($m\ll n$).  \citet{damianou2013deep} leverage the inducing points to train a multi-layer GPs. However, we empirically show that deep GPs with inducing points still suffer from computational challenges in the counterfactual searching phase. 

In this paper, we first link to the previous work\citet{zhang2022interpretable} where supervising the latent space of VAE with probit regression leads to a scalable model for simultaneous classification and counterfactual generation. We demonstrate that the supervised latent space has directed the decoder to generate meaningful counterfactual samples. However, it still remains challenging for the trained decoder to take non-actionable features into account. To improve, we leverage the GPs and adopt the random Fourier features(RFFs) approximations \citet{rahimi} as an alternative to inducing points. The resulting model is equivalent to a two-layer encoder/decoder neural network but requires less trainable parameters. In addition, we construct and learn a density estimator in the latent space, which will be used to generate high-fidelity counterfactual samples. Combining both, we construct a constrained optimization objective for counterfactual searching, allowing us to deploy a mask in the feature space for generating actionable counterfactuals. We demonstrate through various tabular data sets that our model not only achieves comparable performance as a classifier but also illustrates competitive performance under several popular metrics for counterfactuals such as L2 distance, diversity, plausibility, stability, discriminative power, and interpretability. 

\section{Related works}

There are three fields that relate to our work: Gaussian processes in deep learning, counterfactual explanation and generative model.

\subsection{{Gaussian Process in Deep Learning.}} Over the past several decades, researchers have extensively explored the relationship between kernel methods, such as Gaussian Processes (GPs), and neural networks, recognizing their deep theoretical connections in machine learning. Early work by \citet{neal2012bayesian} established the Bayesian interpretation of neural networks, framing them within the probabilistic context that naturally aligns with Gaussian processes. \citet{cho2009kernel} introduced a new family of positive-definite kernel functions called multilayer kernel functions. 

Further advancing this line of research, \citet{lee2017deep} analyzed the behavior of deep neural networks in the infinite-width limit. They demonstrated that as the number of neurons in each layer of a neural network approaches infinity, the network converges to a Gaussian process. This result established an exact equivalence between infinitely wide deep neural networks and Gaussian processes, thereby providing a theoretical foundation for understanding neural networks as a form of Gaussian Process with specific kernel functions. Our proposed model falls within this line of research.

\subsection{Counterfactual explanation}
Counterfactual explanations have a long history in various fields, including social sciences. The concept was first formulated as the optimization problem in Equation \ref{eqn:char_eqn} by \citet{wachter2017counterfactual}.  Following this formulation, numerous studies \citep{grath2018interpretable, dhurandhar2019model, van2019interpretable, dandl2020multi, fernandez2020random, guidotti2018local, mothilal2020explaining} have proposed their own optimization objectives, incorporating different distance metrics and iterative search methods to find counterfactuals. Though effective, the computational cost of the searching algorithms has become the bottleneck of counterfactual generation. To improve, \citet{guo2023counternet,guo2023rocoursenet, zhang2022interpretable} train the encoder as a classifier and the decoder as a counterfactual generator simultaneously with a supervisory signal regularizing the latent space. The counterfactual samples can be generated by linear mapping\cite{zhang2022interpretable} and non-linear mapping\cite{guo2023counternet, guo2023rocoursenet} in the latent space, which effectively reduces the computational cost. Nevertheless, such explainers are model-dependent, and the uncertainty of the decoder still exists. In addition, immutable features still remain challenging to handle under this framework. \citet{guo2023counternet, guo2023rocoursenet} forcibly replace the immutable features with original values, which might result in out-of-distribution samples. In contrast, our approach will be model-agnostic and directly operate in the feature space. 

An alternative approach to generating counterfactuals involves inverse classification, as introduced by \citet{aggarwal2010inverse}. This method seeks to perturb the input data $x$ so that it is classified as a different, counterfactual class. Works such as \citet{lash2017generalized} and \citet{laugel2018comparison} apply inverse classification techniques to produce counterfactual instances, while \citet{sharma2019certifai} specifically aims to alter the input space to flip the classifier’s output to a counter-class. However, a common challenge with these methods is that counterfactual samples might fail to propose meaningful changes despite a change in prediction.

Other researchers have utilized optimization methods guided by a tree structure to search for counterfactuals, as seen in \citet{tolomei2017interpretable}, \citet{fernandez2020random}, and \citet{lucic2019focus}. Additionally, several studies \citep{karimi2020algorithmic, kanamori2020dace, karimi2020model, russell2019efficient, ustun2019actionable} leverage solver-based algorithms to find counterfactuals, particularly for linear or piecewise-linear models. There are also heuristic approaches to solving the optimization problem, explored by \citet{guidotti2018local}, \citet{keane2020good}, \citet{rathi2019generating}, and \citet{white2019measurable}.

While the existing methods have demonstrated efficacy in generating counterfactuals, our approach diverges by employing the Gaussian process with RFFs in an autoencoder framework, resulting in a simpler generative model. Futhermore, we propose a new density estimator to guide the searching process. Our model naturally allows for an immutable attributes by applying a mask. This probabilistic framework aims to enhance the counterfactual generation process by incorporating uncertainty estimation, providing a more robust solution for counterfactual inference.

\subsection{Generative models}
Our work is also related to generative models. In general, there are two popular methods: Generative Adversarial Network (GAN) \cite{goodfellow2014generative} and Variational Auto-Encoder (VAE) \cite{kingma2013auto}. 

Variational Autoencoders (VAEs) have been widely employed as a tool for dimensionality reduction, leveraging their ability to learn compact and informative latent representations. Beyond this, several studies have explored the use of Autoencoders (AEs) and VAEs to generate counterfactual examples \citep{dhurandhar2018explanations, joshi2019towards, van2019interpretable, pawelczyk2020learning, mahajan2019preserving}. Specifically, \citet{pawelczyk2020learning} introduced a VAE-based model called CCHVAE, which is trained on a dataset and utilizes a separate classifier to evaluate the outputs of the decoder. In this approach, a nearest-neighbor search algorithm is employed within the learned latent space to identify counterfactuals. Similarly, \citet{joshi2019towards} implemented a comparable framework, differing primarily in the choice of search algorithm for navigating the latent space. Both methods, however, leave the latent space unsupervised, potentially affecting the quality of the generated counterfactuals.

To address the need for counterfactuals that remain close to the data manifold, \citet{dhurandhar2018explanations} proposed minimizing the L2 distance between a proposed counterfactual and its reconstruction from a pre-trained Autoencoder (AE). \citet{van2019interpretable} combined Autoencoders with k-d trees, utilizing the L2 distance between a sample and its prototype in the latent space to guide the search for counterfactuals. \citet{mahajan2019preserving} addressed the scalability problem by incorporating a hinge loss for classification based on the decoder's output and introducing an oracle to generate feasible counterfactual examples. While this approach alleviates some scalability concerns, it introduces additional computational complexity due to the need for an oracle, which can be resource-intensive.

On the other hand, Generative Adversarial Networks (GANs) have also been explored for counterfactual generation, particularly in the context of image datasets. \citet{samangouei2018explaingan} utilized a GAN model to interpret a classifier's decision on a synthesized image sample. Beyond the standard GAN loss \citep{goodfellow2014generative}, their approach incorporates additional loss terms, including a classification loss, a prior loss, and a reconstruction loss, though it is specifically tailored to image data. More recently, \citet{nemirovsky2020countergan} proposed a GAN-based architecture with a residual connection, allowing for direct generation of counterfactual samples from a given input. This innovative design facilitates more efficient counterfactual generation but is primarily focused on specific data types.

\section{Background}
We begin with a brief review of Gaussian processes (GPs) and the linear approximations to the RBF kernel using random Fourier features (RFF). In the subsequent section, we adapt this approximation for the counterfactual inference task.

Consider a dataset $(x_1, y_1), \dots, (x_n, y_n)$, where $y \in \mathbb{R}$ is an output associated with the input feature vector $x \in \mathbb{R}^d$. Typically, $x$ is known—an assumption we maintain for now, though it will later be treated as a latent embedding. A GP learns the function $y(x)$ according to the following model.

\begin{defn}[Gaussian Process]
For $x \in \mathbb{R}^d$ and $y(x) \in \mathbb{R}$, given a pairwise kernel function $k(x,x')$ between any two points $x$ and $x'$, a Gaussian process is a random function, $y(x) \sim \mathrm{GP}(m(x), k(x,x'))$, such that for any set of $n$ data points $(x_1, y_1), \dots, (x_n, y_n)$, the vector $(y_1, \dots, y_n)$ is Gaussian distributed with mean vector $(m(x_1), \dots, m(x_n))$ and covariance matrix $K_n$ where $K_n(i,j) = k(x_i, x_j)$.
\end{defn}

In this paper, we set $m(x) = 0$ and use the RBF kernel:
\begin{equation}\label{eq.rbfkernel}
k(x,x') = \exp\left\{-\frac{1}{2b}\|x-x'\|^2\right\},
\end{equation}
with parameter $b > 0$. Since $x$ will be learned in our model, allowing the embeddings to scale arbitrarily for any value of $b$, we fix $b=1$ in the following analysis.

A Gaussian process is obtained by integrating over an underlying linear Gaussian model. Let $\phi(x)$ be a mapping of $x$ into another space, and define the linear regression model:
\begin{eqnarray}\label{eqn.linear}
y_i\,|\,x_i, w &\sim& \mathcal{N}(\phi(x_i)^\top w, \sigma^2),\qquad i=1,\dots,n\nonumber\\
w &\sim& \mathcal{N}(0, I).
\end{eqnarray}
The marginal distribution of all $y$ given all $x$ is:
$$y \mid x \sim \mathcal{N}(0, \sigma^2 I_n + K_n),$$
where $K_n(i,j) = \phi(x_i)^\top \phi(x_j)$. Here, $y$ is represented as a noise-added process, but setting $\sigma^2 = 0$ yields the underlying noise-free GP. For the RBF kernel, $\phi(x)$ is a continuous function.

The linear representation in Equation (\ref{eqn.linear}) is preferable for scalability to large datasets. However, because $\phi(x)$ is continuous for the RBF kernel, direct work in this linear space is not feasible. The random Fourier feature (RFF) approach offers an approximation of $\phi(x)$ through Monte Carlo integration as follows.

\begin{defn}[RFF Approximation]
Let $x \in \mathbb{R}^d$ and define a sample size $S$. Generate vectors $z_s \sim \mathcal{N}(0,I)$ in $\mathbb{R}^d$ and scalars $c_s \sim \text{Uniform}(0,2\pi)$ independently for $s=1,\dots,S$. For each $x$, define the vector 
$${\phi}(x) = \sqrt{\frac{2}{S}}\left[\cos\Big(z_1^\top x+c_1\Big),\dots,\cos\Big(z_S^\top x+c_S\Big)\right]^\top.$$
Then ${\phi}(x)^\top{\phi}(x') \rightarrow \exp\{-\frac{1}{2}\|x-x'\|^2\}$ as $S\rightarrow\infty$.
\end{defn}

Using this representation, we revert to the underlying linear model of the Gaussian process by learning a $S$-dimensional model variable $w$ in this new space. Since the approximation to the Gaussian kernel holds, the underlying marginal distribution that this linear model approximates is the desired Gaussian process.

We also observe that the function $\phi(x)$ corresponds to a single layer of a neural network, where the weights $z$ and bias $c$ are random and the nonlinearity is the cosine function. By sampling $z$ and $c$, we reduce the size of trainable parameters and thus prevent it from overfitting. 

\begin{figure*}[t]
\centering
  \includegraphics[width=1\textwidth]{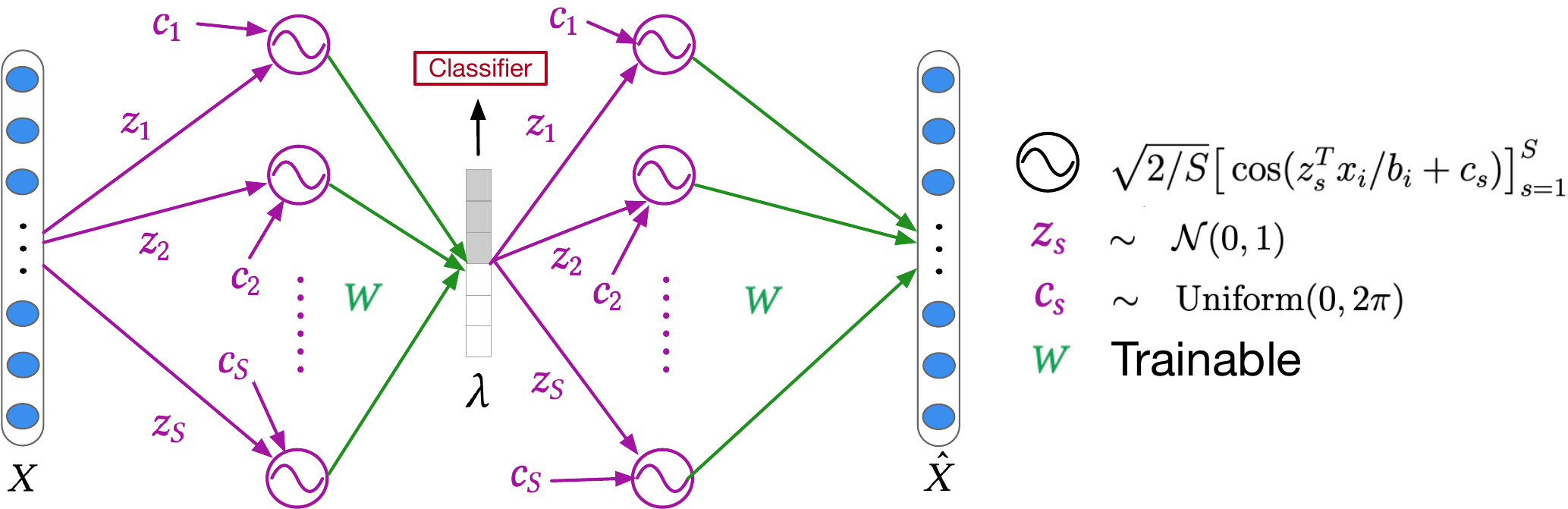}
  \caption{The framework of our GPAE model. The RFF mapping (\textcolor{rff}{purple}) layers is equivalent to a one-layer neural network  with the cosine activations and have fixed weights that do not require training. The only trainable weights(\textcolor{trainable}{green}) are the projection layer in the encoder and the decoder, respectively. The classifier regularizes the latent code, leading to a more interpretable latent space. }
  \label{fig:gpae_framework}
\end{figure*}
\section{Gaussian Process Auto-Encoder (GPAE) and counterfactual generation}

\begin{table}[ht]
\centering
\caption{Definitions of Symbols}\label{table:description}
\begin{tabular}{|c|l|}
\hline
\textbf{Symbol} & \textbf{Description} \\ \hline
$x_i$, $y_i$ & Input data feature vector and label for the $i$-th sample \\ \hline
$f_e(\cdot)$ & Encoder that maps $x_i$ to a latent vector \\ \hline
$f_d(\cdot)$ & Decoder that maps latent vectors to reconstructed data \\ \hline
$\lambda_i^K$ & Latent vector for the $i$-th sample after $K$ flows \\ \hline
$\theta$ & Weight vector in classifier \\ \hline
$\alpha_i$ & Hidden variable for the $i$-th sample in probit model \\ \hline
$\beta_i$ & Projection step size for generating counterfactuals \\ \hline
$w$ & The learnable parameter in density estimator \\ \hline
$W_e$, $W_d$ & The learnable parameter in encoder/decoder \\ \hline
$z_s$, $c_s$ & The fixed parameters in RFF mapping \\ \hline
$\delta$ & The change in the feature sapce \\ \hline
$\gamma$ & The latent vector in $\mathbb{R}^{d-1}$ space \\ \hline
\end{tabular}
\end{table}

In this section, we start with briefly discussing the idea of the supervised auto-encoder(AE) \cite{zhang2022interpretable} in connection to our GPAE model for counterfactual generation. We then propose a density estimator in the latent space for searching a high-fidelity latent vector. With the learned GPAE and density estimator, we then construct a constrained optimization problem that allows us to take immutable features into account while generating meaningful counterfactuals. The description of symbols in this section can be found in Table \ref{table:description}.

Given a data set of $n$ labeled training sample $(x_1,y_1),\dots, (x_n,y_n)$, where $x_i \in \mathbb{R}^D$ is the feature vector and $y_i \in \{-1, 1\}$ is its class label for the $i$-th data point, we aim to learn its latent vector as $\lambda_i \in \mathbb{R}^d$ such that it is not only discriminative but also descriptive for counterfactual generation. Usually $D\gg d$. 

\subsection{Supervised Auto-encoder for counterfactual generation}

\begin{algorithm}[t]
   \caption{SAE with probit model}
   \label{alg:SVAE}
\begin{algorithmic}[1]
   \Require $\mathcal{D} = \{(x_1, y_1),(x_2, y_2),\dots,(x_N, y_N)\}$, $\eta$, $t=0$ \\ 
   \textbf{Initialize} $\zeta_t, \phi_t, \Sigma_t', \mu_t', \mu_{0,t}'$ and $\mu_{\lambda_it},\forall i$ randomly \\
   \textbf{While not converged:} \\ 
    \quad\textbf{Sample} $S_t \sim \mathcal{D}$ \\
    \quad \textbf{Update} $\mu_{\lambda_i}$  and $\mu_t', \Sigma_t', \mu_{0,t}'$ \\
    \quad \textbf{Update} $\zeta_t \gets \zeta_{t-1} + \eta \nabla_{\theta}\mathcal{L}(\zeta, \phi),~~$ $\phi_t \gets \phi_{t-1} + \eta\nabla_{\phi}\mathcal{L}(\zeta, \phi)$ \\
    \textbf{End} \\
    \textbf{return} $\zeta_t, \phi_t, \mu_t', \Sigma_t', \mu_{0, t}'$
\end{algorithmic}
\end{algorithm}

Initially, we incorporate supervised signal in the latent space and regularized the training using probit regression for classification. This approach allows us to learn a more discriminative latent space and use the latent space and decoder to generate a counterfactual sample. Below, we outline the technical details of the supervised Auto-encoder (SAE) in connection with a Gaussian process-based Autoencoder (GPAE). 

The SAE framework consists of an AE architecture: an encoder function, $f_e^{\zeta}(\cdot): \mathbb{R}^D \rightarrow \mathbb{R}^d$, mapping the input data $x_i$ to a latent vector in $\mathbb{R}^d$, and a decoder function, $f_d^{\phi}(\cdot): \mathbb{R}^d \rightarrow \mathbb{R}^D$, which reconstructs the input from its latent representation. The primary objective is to minimize the reconstruction error, ensuring $f_d^{\phi}(f_e^{\zeta}(x)) \approx x$. Both functions are parameterized using neural networks with non-linear activation functions.

We then supervise the AE by incorporating a probit regression model as a binary linear classifier within the AE’s latent space. The embedded probit consists of a weight vector $\theta \in \mathbb{R}^d$, a bias term $\theta_0 \in \mathbb{R}$, and a hidden variable $\alpha_i$ for each data point. The hierarchical structure of the model uses Gaussian priors for $\theta$ and $\theta_0$, defined as:
\[
    \theta \sim \mathcal{N}(0, I), \quad \theta_0 \sim \mathcal{N}(0, \sigma_0^2), \quad \alpha_i \sim \mathcal{N}\big(\langle \lambda_i^K, \theta\rangle + \theta_0, 1\big).
\]
The binary class label $y_i$ is then given by $y_i = \text{sign}(\alpha_i)$. During training, the latent space is guided by the probit model, encouraging a well-defined separation between different class densities, thereby enabling effective counterfactual generation. The algorithm is summarized in Algorithm \ref{alg:SVAE}.

\paragraph{\textbf{SAE with Normalizing Flows (NF)}}
A routine assumption in VAEs is that it has Gaussian prior for latent vector, which might be limited in complex datasets. Normalizing Flows (NF) are applied to enhance the flexibility and expressiveness of the latent space. By applying a series of bijective transformations, the latent space can capture complex data structures beyond traditional Gaussian priors. The overall objective function for SAE with NF is defined as:
\begin{align}
    \mathcal{L}_{\text{SAE}} = \sum\limits_{i=1}^N -\mathbb{E}_{\lambda_i^k}[\log(q(\lambda_i^0))] +  \mathbb{E}_{\lambda_i^k}[\log(p(x_i|\lambda_i^k))] \nonumber\\
    + \mathbb{E}_{\lambda_i^k}[\log(p(\lambda_i^k))] + \sum\limits_{k=1}^K \mathbb{E}_{\lambda_i^k}[\log(q(\lambda_i^k))] \\
    +  \mathbb{E}_{q}\Bigg[\ln\frac{p(y_i, \theta,\theta_0, \alpha, \lambda^K | x)}{q(\theta)q(\theta_0)\prod_{i=1}^Nq(\lambda_i)q(\lambda_i^K|x_i)}\Bigg]
\end{align}
where the expectation is with respect to the chosen $q$ distribution and $K$ is the number of flows. This loss function allows the model to learn richer latent representations while maintaining tractable optimization.

\begin{figure*}[ht]
\centering
  \includegraphics[width=1\textwidth]{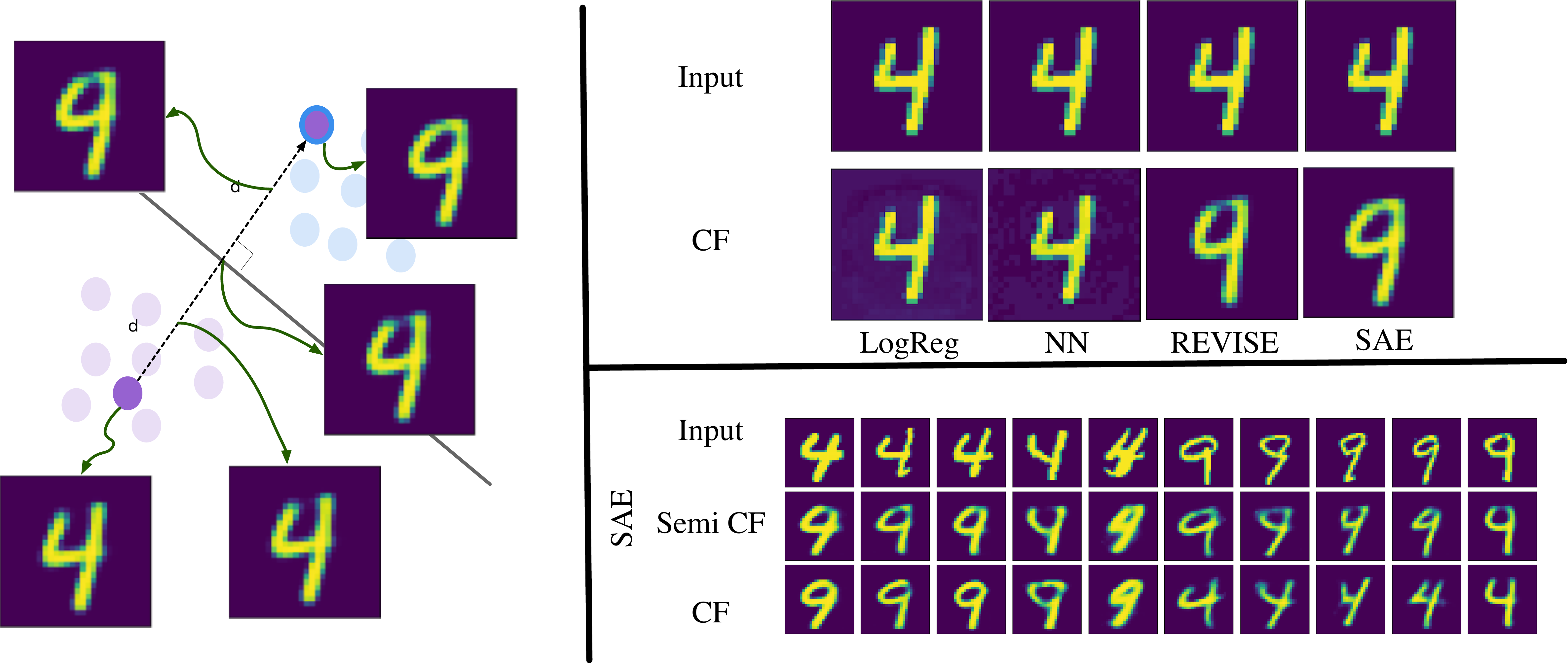}
  \caption{A visual example for illustrative purposes. Left: A rendering of the VAE's latent space and the corresponding decoded images at various locations. Using the decision boundary of a linear classifier in this space, we can generate a sequence of counterfactuals that look increasingly like the counter-class. Top right: Counterfactual examples for the same query input for multiple models (see Table \ref{method_des}). Bottom right: Examples of different factual inputs (row 1), the corresponding counterfactual on the latent decision boundary (row 2) and the CF moving into the counter-class latent space (row 3).}
  \label{fig:mnist_compare}
\end{figure*}

\paragraph{\textbf{Counterfactual Generation}}
Benefiting from the structured latent space induced by the probit model, counterfactuals can be easily generated by projecting a latent vector $\lambda_i^K$ directly onto or beyond the decision boundary. The decision boundary is defined by solving for $\beta_i$ in $(\lambda_i^K + \beta_i \theta)^T \theta + \theta_0 = 0$. The resulting counterfactual latent vector is given by:
\[
    (\lambda_i^K)^c = \lambda_i^K + \text{step} \times \beta_i \theta,\label{eqn:cf_step}
\]
where \textit{step} determines how far to move into the counter-class region. \textit{step}=2 ensures that the generated counterfactuals are meaningful and lie within the counter-class density.

An illustrative example of MNIST dataset is shown in Figure \ref{fig:mnist_compare}. The SAE model contains an autoencoder with Inverse Autoregressive Flow(IAF) \citet{kingma2016improved}. On the left, the counterfactual samples are generated by progressively projecting towards the learned decision boundary and decoding back to the feature space. It can be seen that the learned latent space is meaningful and the projected counterfactual sample makes minimum and necessary changes to flip the class label. On the right, we show different qualitative examples in comparison with other models. 

One potential drawback of SAW is that the decoder, which is a black-box model, might introduce uncertainty while generating counterfactual samples. In addition, it remains challenging to consider immutable features under this framework. Therefore, this motivates our GPAE model. 

\subsection{Gaussian Process Auto-Encoder}
In this subsection, we leverage the Gaussian Process (GPs) to construct our auto-encoder. GPs can be thought of as the approximation of an infinitely wide neural network \cite{lee2017deep}. Motivated by this finding, we construct our AE by using the GPs for the encoder and decoder, respectively. The apparent bottleneck of using GPs is computational cost, which stems from inverting an $NxN$ kernel matrix. The well-studied mitigation introduces inducing points, resulting in a lower rank matrix approximating the full data matrix. The inducing points can also be learned using the variational objective by maximizing the ELBO. However, as we show later in the experimental section, the computational burden is still high in producing counterfactual samples in our framework and there is little performance improvement. 

We approach the problem with Random Fourier features approximation and effectively reduce the computational burden in our framework. The resulting model is equivalent to an AE with two-layer neural networks, achieving competitive performance at a higher speed. We let $f_e(\cdot)$ be an encoder and $f_d(\cdot)$ be an decoder. In our scenario, each encoder or decoder is a Gaussian process with the Random Fourier features approximation followed by a linear mapping. 

In particular, we define the encoder to be the random function $f_e(x_i) \sim GP(0,k(x_i,x_j))$ and use the RFF approximation to learn it. Thus, the encoder becomes:
\begin{eqnarray}
    f_e(x_i) &=& \phi(x_i)^\top W_e,\label{eqn.fe}\nonumber\\
    W_e &\sim & \cN(0,1)^{S\times d},
\end{eqnarray}
where $W_e\in\mathcal{R}^{S\times d}$ and $\phi(x_i) \in [-\sqrt{2/S},\sqrt{2/S}]^S$ is the RFF map:
\begin{eqnarray}
    \phi(x_i) &=& \sqrt{2/S} \big[\cos(z_s^T x_i/b_i+c_s)\big]_{s=1}^S,\label{eqn.phi}\nonumber\\
    z_s &\sim & \cN(0,I) \quad z_s\in\mathcal{R}^d,\\
    c_s &\sim & \mathrm{Uniform}(0,2\pi).\nonumber    
\end{eqnarray}

Let $\lambda_i = f_e(x_i)$ be the encoded latent vector for i-th sample. Similarly, our decoder resembles the encoder form with the same RFF mapping:
\begin{eqnarray}
    f_d(\lambda_i) &=& \phi(\lambda_i)^\top W_d,\label{eqn.fd}\nonumber\\
    W_d &\sim & \cN(0,1)^{S\times D},
\end{eqnarray}
where $W_d\in\mathcal{R}^{S\times D}$ and $\phi(\lambda_i) \in [-\sqrt{2/S},\sqrt{2/S}]^S$ is the same RFF mapping as Equation \ref{eqn.phi}.

We define $\hat{x}_i = f_d(\lambda_i)$ be the reconstructed sample. We apply the squared error between $x_i$ and $\hat{x}_i$. Then, the loss for all the training data can be computed as:
\begin{eqnarray}
\mathcal{L}_{recon} = \frac{1}{n} \sum\limits_{i=1}^n ||x_i - \hat{x}_i ||^2
\end{eqnarray}

\begin{algorithm}[t]
\caption{GPAE for counterfactual inference}\label{alg_gpae}
\begin{algorithmic}[1]
\Require Data $\{(x_1,y_1),(x_2,y_2),\dots,(x_n,y_n)\}$, GP width $S$, kernel widths $b$.\\
 \textbf{Sample} $z_s^e,z_s^d \sim \cN(0,I)$, $c_s^e, c_s^d \sim \textrm{Unif}(0,2\pi)$, $s = 1:S$.\\ \qquad Alternatively, grid using inverse CDF.\\
 
 \textbf{Define} $\phi(x_i) = \sqrt{2/S} \big[\cos((z_s^e)^T x_i/b_i+c_s^e)\big]_{s=1}^S$\\
 \textbf{Define} $\phi(\lambda_i) = \sqrt{2/S} \big[\cos((z_s^d)^T \lambda_i/b_i+c_s^d)\big]_{s=1}^S$\\
 
 \textbf{While not converged:}\\
 \quad\textbf{Compute} $\lambda_i = \phi(x_i)^TW_e$ \\
 \quad\textbf{Compute} $\hat{x}_i = \phi(\lambda_i)^TW_d$ \\
 \quad\textbf{Compute} $\hat{y}_i = sigmoid(\lambda_i^T \theta_c$) \\
 \quad\textbf{Gradient Descent on $W_e, W_d, \theta_c$}\\
 \textbf{Return $W_e, W_d, \theta_c$}
\end{algorithmic}
\end{algorithm}

\subsection{Classifier}
We construct a linear classifier in the latent space for a binary prediction. Then, the logits of the classifier can be computed as:
\begin{eqnarray}
    f_c(\lambda_i) = \lambda_i^T \theta_c
\end{eqnarray}
where $\theta_c$ is a learnable parameter of the decision boundary. 

We then use the sigmoid function to compute the class probability for each sample. We apply a binary cross-entropy loss for this task as:
\begin{eqnarray}
    \mathcal{L}_{class} = -\frac{1}{n} \sum\limits_{i=1}^n y_i\log(\hat{y}_i)
\end{eqnarray}

The pipeline of our model is shown in Figure \ref{fig:gpae_framework}.

\subsection{Density Estimation}

\begin{figure}
\centering    
\subfigure[]{\label{fig:dens_scatter}\includegraphics[width=60mm]{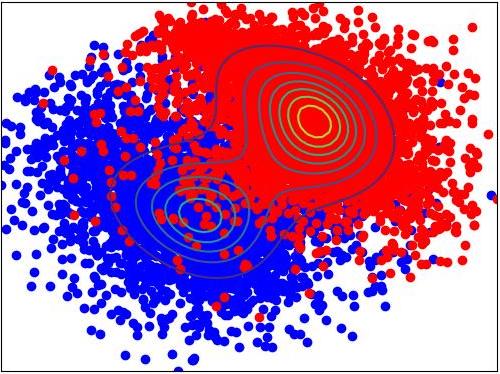}}
\subfigure[]{\label{fig:density_contour}\includegraphics[width=60mm]{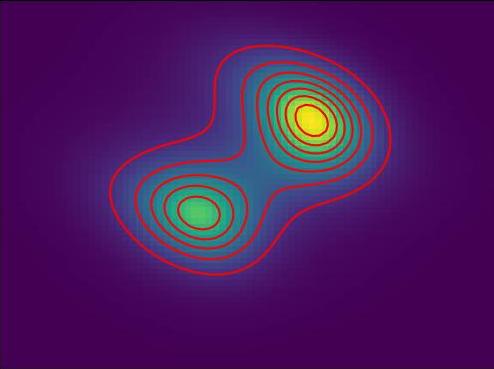}}
\caption{The 2D visualization of density estimation of LCD dataset. (a)The scatter plot of LCD data samples in 2-dimensional latent space and the contour level sets. (b) Heatmap of the estimated density of (a) and its corresponding contour level sets. }\label{fig:density}
\end{figure}

The counterfactual samples generally require changing their predicted label with minimal changes in the feature space. We aim for the counterfactuals that not only bear minimal changes but also stay in the data distribution. Before we discuss our counterfactual searching algorithm, we first construct our density estimator in the latent space. Let $\lambda_1, \lambda_2, \dots, \lambda_N$ be the N observed data samples and $\lambda_i \in \mathcal{R}^d$. We also assume a random function over $\lambda$, $f(\lambda) \sim \mathcal{GP}(0,K)$ with some kernel $K$. Then, we construct our density estimator as:

\begin{eqnarray}
    p(\lambda) = \frac{e^{f(\lambda)}\mathcal{N}(\lambda|\mu,\sigma)}{\int e^{f(\lambda)}\mathcal{N}(\lambda|\mu,\sigma) d\lambda}
\end{eqnarray}

In the numerator, the random function $f(\lambda)$ is first exponentiated to stay positive, and then the Gaussian envelope is applied to trim the tails. Here, we again use RFF approximation over the random function $f(\lambda)$, which results in the simpler form $f(\lambda) = w^T\phi(\lambda)$ where $\phi(\lambda)$ is the RFF mapping and $w\sim\mathcal{N}(0, I)$. The denominator is the normalizing constant. 

We learn our density estimator using Maximum a Posteriori objective to find the optimal $w$. The objective is derived as:
\begin{eqnarray}
    \arg\max_w \mathcal{L}(w) &=&\log\bigg( p(w)\prod_{i=1}^N p(\lambda_i|w)\bigg) \\
    &\approx& -\frac{1}{2} w^Tw + \bigg[\sum\limits_{i=1}^N w^T\phi(\lambda_i)\bigg] - N\log\frac{1}{K}\sum\limits_{k=1}^K e^{w^T\phi(\lambda_k)} 
\end{eqnarray}
Taking the derivative with respect to $w$ leads to:
\begin{eqnarray}
    \frac{1}{N}\frac{\partial \mathcal{L}}{\partial w} &=& -\frac{w}{N} + \frac{1}{N}\sum\limits_{i=1}^N \phi(\lambda_i) - \sum\limits_{k=1}^K\frac{ e^{w^T\phi(\lambda_k)} }{\sum\limits_{j=1}^K e^{w^T\phi(\lambda_j)}}\phi(\lambda_k) \\
    &=& -\frac{w}{N} + \frac{1}{N}\sum\limits_{i=1}^N \phi(\lambda_i) - \sum\limits_{k=1}^K q(\lambda_k) \phi(\lambda_k)
\end{eqnarray}
where $q(\lambda_k) = e^{w^T\phi(\lambda_k)}/\sum_{j=1}^K e^{w^T\phi(\lambda_j)}$. We apply gradient descent to learn the density estimator $w$. The demonstration of the learned density estimator from LCD data set is shown in Figure \ref{fig:density}. In the figure, the scatter plot shows the empirical distribution of data samples in 2D latent space. The heatmap on the right illustrates the estimated density from the samples. It is evident that the estimated density highly aligns with the empirical distribution, which plays a crucial role in finding a meaningful counterfactual sample in the searching algorithm that we discuss next. 

\subsection{Counterfactual generation using density estimation}
Once we learn the density estimator, we can now construct our counterfactual searching algorithm. Let $x$ be a query sample. Its counterfactual can be generated by $x^c = x + \delta$ where $\delta$ is the proposed change in the feature space and $\lambda$ is the latent code for $x$. Therefore, $\lambda^c = f_e(x^c) = f_e(x + \delta) $. To find the counterfactual for a query sample, we search for a sample such that it has the minimum $\delta$, lies on the decision boundary and stays in the high-density area in the latent space. These constraints lead us to optimize:
\begin{eqnarray}
    \min_{\delta} &&\frac{1}{2}||\delta ||^2 - \beta \log\big(p(f_e(x + \delta))\big) \\
          s.t. && \theta_c f_e(x + \delta)  = 0
\end{eqnarray}
where $\beta$ is the regularizing factor on the log density. The benefit of this optimization objective is that we force the counterfactual to stay on the decision boundary, which naturally makes the counterfactual sample valid (i.e., changes its predicted label to the target class) by this setup.

The Lagrangian function of the above constraint problem can be transformed into:
\begin{eqnarray}
    \mathcal{L}(\delta, \eta) &=& \frac{1}{2}||\delta ||^2 - \beta \log\big(p(f_e(x + \delta))\big) + \eta (\theta^T \lambda^c + \theta_0)  \\
    &=& \frac{1}{2}\delta^T\delta  \nonumber\\
    & & - \beta\bigg[w^T\phi(f_e(x+\delta)) - \frac{1}{2}\big(f_e(x+\delta)^T\Sigma^{-1}f_e(x+\delta)) + \mu^T\Sigma^{-1}f_e(x+\delta)\bigg]  \nonumber \\
    & & + \eta\theta^Tf_e(x+\delta) + \eta\theta_0 + const.
\end{eqnarray}
where $\eta$ is the Lagrangian multiplier. The corresponding dual function is to:
\begin{eqnarray}
    \max_{\eta}\min_{\delta} \mathcal{L}(\delta, \eta)
\end{eqnarray}
which require the derivative of both parameters. First, taking the derivative with respect to $\delta$ leads to:
\begin{eqnarray}
    \frac{\partial \mathcal{L}}{\partial \delta} &=& \delta - \nonumber \\
    & & \beta\bigg[-\sqrt{\frac{2}{S}}\nabla_{\delta}f_e(\delta)[(sin(Z\lambda^c + C)\odot Z)^T w ]\nonumber \\
    & &- \nabla_{\delta}f_e(x+\delta)[f_e(x+\delta)^T\Sigma^{-1}] + 2\nabla_{\delta}f_e(x+\delta)[\mu^T\Sigma^{-1}]\bigg] \nonumber \\
    & & + \eta\nabla_{\delta}f_e(x + \delta)\theta
\end{eqnarray}
where $\nabla_{\delta}f_e(x^c)$ can be calculated:
\begin{eqnarray}
    \nabla_{\delta}f_e(x+\delta) = -\sqrt{\frac{2}{S}}(sin(Z(x+\delta) + C)\odot Z)^Tw_e
\end{eqnarray}
Secondly, the derivative with respect to $\eta$ simply leads to:
\begin{eqnarray}
    \frac{\partial \mathcal{L}}{\partial \eta} = \theta^Tf_e(x^c) + \theta_0
\end{eqnarray}
The optimal $\delta$ can be found by iteratively and interchangeably optimizing both parameters. 

\paragraph{\textbf{Masked counterfactual}} One of the benefits is that this setup allows the counterfactual sample to be constrained when a subset of features is immutable. In this scenario, we introduce a mask $M\in\{0,1\}^D$ where $0$ means changes are not allowed and $1$ means otherwise. Then, we apply the mask in the input space to obtain the final update for $\delta$:
\begin{eqnarray}
    \delta_{t+1} = \delta_t - step*\frac{\partial \mathcal{L}}{\partial \delta}\bigg|_{\delta = \delta_t} \odot M
\end{eqnarray}

\subsection{Selection of $\beta$}\label{sec:select_beta}
\begin{figure*}[t]
\centering
  \includegraphics[width=1\textwidth]{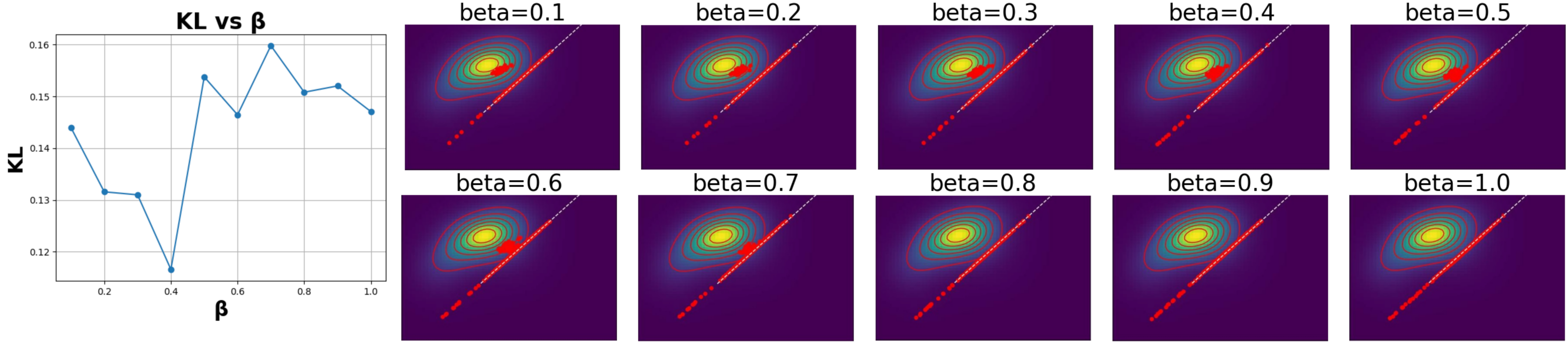}
  \caption{The study on the selection of $\beta$ in LCD data set. Left: the KL divergence between true density and estimated density in latent space vs $\beta$. The KL is minimized around $\beta=0.4$. The KL increases with lower or higher $\beta$, indicating less matching between the estimated density and true density of the learned latent representation. Right: the visualization of 2D latent space of the target class (upper left corner) and counterfactual samples (from rejected class). As $\beta$ increases, the latent counterfactuals tend to stay closer. }
  \label{fig:latent_KL_beta}
\end{figure*}

We notice that the regularizing parameter $\beta$ significantly affects the quality of counterfactual samples. The larger $\beta$ it is, the more concentrated the counterfactual sample will be around the decision boundary. From the latent space view, as the $\beta$ increases, the latent vectors tend to squeeze more into the dense area on the decision boundary. This is illustrated in the right plot in Figure \ref{fig:latent_KL_beta}. Although it is favorable to stay in the high-density area, it can lead to less diversified counterfactual generation. Performing grid search on $\beta$ is infeasible due to the high computational cost. Thus, this motivates an algorithm for selecting an appropriate $\beta$ by comparing the distance between the counterfactual distribution and the projected data distribution on the decision boundary. 

In particular, we search $\beta$ by measuring the KL divergence between the approximate distribution and true distribution of the latent vectors on the decision boundary. This requires us to project all the vectors onto the decision boundary. Let $\lambda \in \mathcal{R}^d$ denote a latent vector and $p(\lambda | w)$ be the probability density in the latent space. Given the learned decision boundary $\theta_c$, we construct the projection matrix that maps each latent code onto the decision boundary. let's define the basis vectors in the latent space as:
\begin{eqnarray}
   B_{new} = \left[\begin{array}{c|c|c|c}
                    \vdots & \vdots & \cdots & \vdots\\
                    b_1 = \theta_c & b_2 & \cdots & b_d \\
                    \vdots & \vdots & \cdots & \vdots  \\
                \end{array}\right ]
\end{eqnarray}
The first basis vector is the learned decision boundary $\theta_c$. Then, the rest of the basis vectors can be found using Gram-Schmidt process so that every pair of the basis vectors is orthonormal. 

Once the Gram-Schmidt process is complete, the projection matrix $A \in \mathcal{R}^{d\times d-1}$ is just:
\begin{eqnarray}
   A = \left[\begin{array}{c|c|c|c}
                    \vdots & \vdots & \cdots & \vdots\\
                    b_2 & b_3 & \cdots & b_d \\
                    \vdots & \vdots & \cdots & \vdots  \\
                \end{array}\right ]
\end{eqnarray}

We can apply the project matrix on the $\lambda \in \mathcal{R}^d$ to obtain the projected latent vector $\gamma \in \mathcal{R}^{d-1}$ on the decision boundary. Our density estimator with respect to $\gamma$ is defined as:
\begin{eqnarray}
    p(\gamma) = \frac{e^{w^T \phi(A\gamma)}\mathcal{N}(A\gamma|\mu,\sigma)}{\int e^{w^T \phi(A\gamma)}\mathcal{N}(A\gamma|\mu,\sigma) d\gamma}
\end{eqnarray}

Empirically, we model each latent vector on the decision boundary as a Gaussian random variable with projected mean and fixed variance. We know that the transformation $\lambda_n^c = A\gamma$ leads to $\gamma = A^T\lambda_n^c$. The empirical distribution over $N$ data samples on the decision boundary can be calculated as:
\begin{eqnarray}
    q(\gamma) = \frac{1}{N}\sum\limits_{n=1}^N \mathcal{N}(\gamma|A^T\lambda_n^c, \Sigma)
\end{eqnarray}
where $\Sigma = \sigma^2I_{d-1}$. Then the KL divergence can be measured between q and p distributions in the latent space as:
\begin{eqnarray}
    KL(q(\gamma)|p(\gamma)) &=& \int q(\gamma)\ln \frac{q(\gamma)}{p(\gamma)} d\gamma \\
    &=& \mathbb{E}_q\bigg[\ln \frac{\frac{1}{N}\sum\limits_{n=1}^N \mathcal{N}(\gamma|A^T\lambda_n^c, \Sigma)}{e^{w^T \phi(A\gamma)}\mathcal{N}(A\gamma|\mu,\sigma)}\bigg] + const.
\end{eqnarray}
We can approximate the KL divergence using the Monte Carlo sampling method. We generate L samples from the q distribution and calculate the mean of $\ln(q/p)$. The algorithm is summarized in Algorithm \ref{alg_mc_kl}. 

\begin{algorithm}[t]
\caption{Monte Carlo estimation of KL divergence}\label{alg_mc_kl}
\begin{algorithmic}[1]
\Require $p(\gamma)$, $q(\gamma)$, number of steps $L$, number of samples $N$\\
 \textbf{Sample} $\gamma_i^l \sim q(\gamma), \gamma_i \in\mathcal{R}^{L\times d-1}, i\in [1,2,\cdots,N]$\\
 
 \textbf{Define} total=0\\
 
 \textbf{for L iterations:}\\
 \quad\textbf{Compute} $\gamma^l = \frac{1}{N}\sum\limits \gamma_i^l$ \\
 \quad\textbf{Compute} total $+=\ln(q(\gamma^l)/p(\gamma^l))$ \\
 \textbf{Return total/L}
\end{algorithmic}
\end{algorithm}

The demonstration of the selection of $\beta$ on LCD data is shown in Fgiure \ref{fig:latent_KL_beta}. On the left, we plot the KL divergence versus different $\beta$ from 0.1 to 1.0. As the $\beta$ increases, the trend of KL divergence first decreases and then increases, implying that a higher $\beta$, though favorable in terms of a higher density score, leads to less density matching to the existing data samples. On the right, we visualize the effect of a higher $\beta$ in 2D space. 

\section{Empirical studies}
We concentrate on five large-scale tabular datasets related to credit lending and law school applications. We All experiments were implemented using PyTorch.

\subsection{Data sets}
We perform experiments on several large-scale tabular data sets frequently used for counterfactual tasks. This includes \textbf{FICO}\footnote{\url{https://community.fico.com/s/
explainable-machine-learning-challenge}}, \textbf{Adult}\footnote{\url{https://www.kaggle.com/datasets/wenruliu/adult-income-dataset}}, \textbf{LAW}\footnote{\url{https://www.kaggle.com/datasets/danofer/law-school-admissions-bar-passage/code}} tabular data sets. For these, we follow the processing in \citet{chang2021node,popov2019neural}. We also consider our own processing of credit lending data sets \textbf{LCD}\footnote{\label{github} \url{https://github.com/Wei2624/GPNAM}} and \textbf{GMC}\footref{github} More information about these data sets is shown in Table \ref{table:datasets}.

\begin{table}[t]
\centering 
\begin{tabular}{ c c c c c  }
 \hline
 \textbf{Dataset} & \textbf{$\#$Train} & \textbf{$\#$Val} & \textbf{$\#$Test} & \textbf{$\#$Feat}\\ 
 \hline
  FICO            & 7,321  & 1,000 & 1,000 & 39 \\
  LCD             & 10,000 & 1,000 & 1,000 & 7 \\
  GMC             & 15,000 & 1,000 & 1,000 & 10  \\
  LAW             & 5,502 & 1,000 & 1,000 & 11  \\
  Adult           & 47,842 & 1,000 &1,000  & 11\\
 \hline
\end{tabular}
\caption{Statistics from the tabular data sets used in counterfactual generation.}\label{table:datasets}
\end{table}

\subsection{Methods and evaluation}
We evaluate our method by comparing it with several existing counterfactual approaches, as summarized in Table \ref{method_des}. A brief description of these methods is provided below.

As a baseline, we use logistic regression for both classification and counterfactual generation. Logistic regression learns a weight vector for the input space, allowing counterfactual samples to be generated by projecting an input directly using the learned weight vector, which remains consistent across all inputs. We also compare the classification performance with a neural network consisting of an architecture identical to the encoder of our GPAE model. In this scenario, counterfactual samples are generated by iteratively optimizing the input space until the model changes the predicted label to the counter-class, following the approach proposed by \citet{wachter2017counterfactual}, specifically $x' = x + \partial f(x)/\partial x$, where $f$ is the deep classifier. Additionally, we benchmark our method against popular counterfactual models that leverage VAEs for counterfactual generation \cite{pawelczyk2020learning, joshi2019towards, downs2020cruds, antoran2020getting}. In all cases, the model being explained is identical to the encoder of our GPAE model. We implement these benchmarking methods using the CARLA library \cite{pawelczyk2021carla}, employing the default parameters for all methods.

We also compare our GPAE with the GP that often utilizes inducing points to reduce the computational cost. Similar to our architecture, we introduce a 2-layer deep GPs as the encoder and decoder, respectively. The benefit of GP with inducing points allows us to choose a wide range of kernel options such as Automatic Relevance Determination (ARD) and RBF kernel. The computational cost of these kernels can be reduced by learning inducing points. The inducing points are pseudo points that are able to approximate the full rank of the covariance kernel matrix and are learned in the variational setting. The inducing points enable the deep GP to infer faster by avoiding inverting the full data matrix. However, we notice that even if inducing points are applied, it is significantly slower than our proposed RFF GP while performing counterfactual searching. Our GPAE still outperforms in nearly all metrics. In all experiments, we select 128 inducting points. 

Next, we introduce several widely used metrics to evaluate the performance. The detailed description of each metric is as follows:

\begin{table}[t]
\centering
\caption{Methods evaluated in our experiments. Cited methods are learned using the CARLA library \cite{pawelczyk2021carla}.} 
\begin{tabular}{ c|c}
\hline
\textbf{Notation} & \textbf{Description} \\ 
\hline
  LogReg    & Binary logistic regression model\\
  Wachter   & \citet{wachter2017counterfactual} \\
  CCHVAE    & \citet{pawelczyk2020learning} \\
  REVISE    & \citet{joshi2019towards} \\
  CRUDS     & \citet{downs2020cruds}\\
  CLUE      & \citet{antoran2020getting} \\
  FACE      & \citet{poyiadzi2020face} \\
  SAE       & \citet{zhang2022interpretable} \\
  CounterNet& \citet{guo2023counternet} \\
  GPAE-ind  & Our model with inducing points\\
  GPAE      & Our model with RFF mapping\\
\hline
\end{tabular}
\label{method_des}
\end{table}

\paragraph{\textbf{Euclidean Distance}} The Euclidean distance measures how much changes have been made in the counterfactual sample from the query sample. We use L2 distance defined as:
\begin{equation}
    \text{L2} = \textstyle \frac{1}{N}\sum\limits_{i=1}^N ||x_i - x_i^{cf}||_2^2
\end{equation}
Note that L2 distance is not defined for categorical features. As such, we only evaluate on continuous features. In addition, we also notice that a lower L2 distance might not necessarily mean a preferable interoperability score. We define the additional metrics below to prove more comprehensive evaluations on the generated counterfactual samples. 

\paragraph{\textbf{Diversity.}} Diversity measures the variety within the group. It provides additional performance information because low IM1 and IM2 may occur with counterfactuals that tend to merge to a single point; not only should the counterfactual look like the counterclass, it should also preserve its variety. The diversity metric is calculated as
\begin{equation}
    \text{Diversity} = \textstyle \frac{1}{N(N-1)}\sum_{i=1}^N\sum_{j=i+1}^N d(x_i^{\mathrm{cf}},x_j^{\mathrm{cf}})
\end{equation}
where $d(\cdot, \cdot)$ is a predefined distance function. We use the Euclidean distance in this paper. 

\paragraph{\textbf{Instability.}} A stable counterfactual explainer should produce similar counterfactual outputs for two similar query inputs. Instability quantifies this as
\[
\text{Instability} = \textstyle \frac{1}{N} \sum_{i=1}^N \frac{d(x_i^{\mathrm{cf}},\widehat{x}_i^{\mathrm{cf}})}{1 + d(x_i,\widehat{x}_i)}
\]
where $\widehat{x}_i = \arg\min_{x\in X\setminus x_i, f(x) = f(x_i)} \|x-x_i\|$, the point within the data set closest to $x_i$ that has the same label. A low instability is preferred.

\paragraph{\textbf{Discriminative power.}} As discussed in \citet{mothilal2020explaining}, counterfactual samples should be highly discriminative to reveal the reasons for the classifier's decision. The determination is usually subject to human judgment, which is costly. An alternative approach is to train a 1-NN classifier based on $x$ and its counterfactual $x^{cf}$. We then select $k$ closest samples from $X_{=}\in X$ and $X_{\neq} \in X$ where $f(X_{=}) = f(x)$ and $f(X_{\neq}) \neq f(x)$. The discriminative power is defined as the classification accuracy of 1-NN on the $X_{=} \cup X_{\neq}$. The higher, the better. 

\paragraph{\textbf{Interpretability.}} \citet{van2021interpretable} use an autoencoder to evaluate the \textit{interpretability} of a counterfactual method. Let $\mathrm{AE}_o$, $\mathrm{AE}_t$, and $\mathrm{AE}$ be three autoencoders trained on the original class, target class, and the entire dataset, respectively. The IM1 and IM2 scores are
\begin{align}
    \mathrm{IM1} &= \textstyle \frac{1}{N}\sum_{i=1}^N\frac{\|x_i^{\mathrm{cf}} - \mathrm{AE}_t(x_i^{\mathrm{cf}})\|^2}{\|x_i^{\mathrm{cf}} - \mathrm{AE}_o(x_i^{\mathrm{cf}})\|^2 + \epsilon} \nonumber \\
    \mathrm{IM2} & =  \textstyle \frac{1}{N}\sum_{i=1}^N\frac{\|\mathrm{AE}_t(x_i^{\mathrm{cf}}) - \mathrm{AE}(x_i^{\mathrm{cf}})\|^2}{\|x_i^{\mathrm{cf}}\|_1 + \epsilon}
\end{align}
where $x_i^{\mathrm{cf}}$ is the $i$th of $N$ counterfactuals. Lower value of IM1 indicates that the generated counterfactuals are reconstructed better by the autoencoder trained on the counterfactual class ($\mathrm{AE}_t$) than the autoencoder trained on the original class. This suggests that the counterfactual is closer to the data manifold of the counterfactual class, and thus more plausible. A similarly interpretation holds for IM2. Hence, lower values of IM1 and IM2 are preferred.



\paragraph{\textbf{Validity.}} This metric verifies that the generated counterfactual indeed lies in the counter-class region of the classifier to be explained. This is
\begin{equation}
    \text{Validity} = \textstyle \frac{1}{N}\sum_{i=1}^N \mathbbm{1}(f(x_i^{\mathrm{cf}}) = y')
\end{equation}
where $f(\cdot)$ is the explained classifier and $y'$ is the target label. (Not all counterfactual methods generate counterfactuals that are guaranteed to change their label.)

For all experiments, the training batch size for each update is 512 random samples from the training data sets. The initial learning rate is $10^{-3}$ and decreases by 10 times after the loss fails to significantly decrease for 10 consecutive epochs. The training stops after the loss fails to significantly decrease for 20 consecutive epochs. We choose $\beta$ following the algorithm in Section \ref{sec:select_beta}.

\begin{longtable}{c c|c|c|c|c|c }
\hline\hline
\multicolumn{7}{c}{\textbf{Classification performance(\%)}}\\
\hline
\multicolumn{2}{c|}{\textbf{Model}} & LCD & GMC & FICO & Adult & LAW \\
\hline
\multirow{4}{*}{\textbf{LogReg}} 
& Accu. & 90.1 & 69.3 & 71.6 & 67.8 & 77.4\\
& Prec. & 87.7 & 65.7 & 71.7 & 88.0 & 71.8\\
& Rec.  & 93.2 & 80.4 & 75.7 & 41.4 & 90.2\\
& AUC   & 93.2 & 74.5 & 72.2 & 85.0 & 87.8\\
\hline
\multirow{4}{*}{\textbf{SVM}} 
& Accu. & 90.4 & 74.5 & 71.2 & 72.9 & 77.1\\
& Prec. & 86.9 & 68.5 & 70.5 & 91.2 & 71.1\\
& Rec.  & 95.0 & 90.6 & 77.4 & 50.8 & 91.0\\
& AUC   & 94.4 & 81.2 & 78.4 & 90.1 & 86.2\\
\hline
\multirow{4}{*}{\textbf{Decision Tree}} 
& Accu. & 83.9 & 70.5 & 64.0 & 75.2 & 78.1\\
& Prec. & 83.5 & 70.5 & 65.2 & 84.3 & 73.2\\
& Rec.  & 82.2 & 70.4 & 66.8 & 62.4 & 89.0\\
& AUC   & 83.9 & 70.5 & 64.4 & 78.5 & 85.2 \\
\hline
\multirow{4}{*}{\textbf{Adaboost}} 
& Accu. & 89.2 & 76.3 & 72.4 & 76.4 & 77.4 \\
& Prec. & 86.1 & 72.1 & 72.7 & 90.7 & 72.1 \\
& Rec.  & 93.4 & 85.8 & 75.7 & 58.8 & 89.2\\
& AUC   & 95.0 & 81.3 & 79.3 & 91.0 & 88.1\\
\hline
\multirow{4}{*}{\textbf{Random Forest}} 
& Accu. & 89.3 & 76.6 & 70.2 & 67.2 & 75.6\\
& Prec. & 86.7 & 73.1 & 69.1 & 93.3 & 72.1\\
& Rec.  & 92.8 & 84.0 & 78.1 & 37.0 & 89.2\\
& AUC   & 92.8 & 84.0 & 76.4 & 89.8 & 88.1\\
\hline
\multirow{4}{*}{\textbf{Gaussian NB}} 
& Accu. & 89.4 & 71.7 & 62.5 & 64.0 & 79.3 \\
& Prec. & 86.9 & 70.8 & 72.1 & 88.0 & 78.4 \\
& Rec.  & 92.9 & 87.4 & 48.1 & 32.4 & 80.8\\
& AUC   & 95.0 & 82.0 & 73.1 & 85.2 & 86.3\\
\hline
\multirow{4}{*}{\textbf{Neural Networks}} 
& Accu. & 89.1 & 77.6 & 71.9 & 65.2 & 76.3 \\
& Prec. & 87.7 & 74.5 & 71.6 & 88.0 & 70.0 \\
& Rec.  & 91.3 & 85.2 & 76.7 & 35.2 & 92.0 \\
& AUC   & 95.1 & 85.9 & 78.7 & 86.0 & 87.7 \\
\hline
\multirow{4}{*}{\textbf{SAE}} 
& Accu. & 89.1 & 76.5 & 72.8 & 78.1 & 79.1\\
& Prec. & 87.5 & 70.9 & 74.1 & 86.9 & 73.8\\
& Rec.  & 91.2 & 90.0 & 70.1 & 63.2 & 90.1\\
& AUC   & 95.2 & 85.7 & 80.4 & 80.2 & 79.4\\
\hline
\multirow{4}{*}{\textbf{GPAE-ind}} 
& Accu. & 90.3 & 75.6 & 65.6 & 71.0 & 74.1\\
& Prec. & 82.2 & 68.3 & 63.2 & 85.3 & 68.0\\
& Rec.  & 96.0 & 88.0 & 74.7 & 59.8 & 91.2\\
& AUC   & 91.3 & 79.6 & 69.8 & 72.0 & 75.1\\
\hline
\multirow{4}{*}{\textbf{GPAE}} 
& Accu. & 90.1 & 76.9 & 72.0 & 77.4 & 78.7 \\
& Prec. & 87.0 & 73.8 & 72.3 & 86.4 & 73.5 \\
& Rec.  & 94.4 & 83.4 & 75.2 & 66.5 & 89.7 \\
& AUC   & 95.3 & 82.5 & 79.0 & 77.4 & 78.7\\
\hline
\caption{Classification results of accuracy, precision, recall and Area Under Curve(AUC) across different data sets. }
\label{tab:clf_results}
\end{longtable}

\subsection{Classification performance}
Although our goal is not to construct the state-of-the-art classifier, we also study the classification performance across all the data sets. We aim to show that 1) our encoder itself as a classifier can achieve competitive performance in classification with other popular classifiers, especially a neural network, and 2) the learned classifier in our framework deserves to be explained. After all, it is not worth explaining a bad classifier. We compare our model with other typical types of classifiers, including neural networks, which consist of the same architecture as our encoder. Our auto-encoder architecture handles two tasks simultaneously during training, the generative task and the classification task. The competitive classification performance implies that the learned decision boundary in the latent space is meaningful and distinguishable. We compare accuracy, precision, recall and AUC for each model. The results are shown in Table \ref{tab:clf_results}. 

Overall, the results show that our proposed model achieves a competitive classification performance to that of other benchmark models. In particular, we find that the GPAE model occasionally outperforms neural networks on certain datasets, such as LCD and FICO, in terms of AUC, indicating that the learned decision boundary effectively separates data points in the latent space irrespective of the threshold. Despite maintaining competitive performance for accuracy, precision, and recall on the LAW and Adult datasets, we note that the GPAE model does not achieve a similar level of AUC. The model appears to be sensitive to classifier thresholds, likely due to training on imbalanced datasets. As we are not targeting to design a state-of-the-art classifier, the competitive performance on precision and recall still allows us to explain the classifier at a particular threshold. 

\subsection{Counterfactual evaluation}
\begingroup
\setlength{\tabcolsep}{4pt} 
\scriptsize
\begin{longtable}{c c| c|c|c|c|c|c|c }
\hline\hline
\multicolumn{9}{c}{\textbf{Counterfactual Evaluations}}\\
\hline
\multicolumn{2}{c|}{\textbf{Model}}& L2$\downarrow$ & Div.$\uparrow$  & Instb.$\downarrow$  & Dispo.$\uparrow$  & IM1$\downarrow$  &  IM2$\downarrow$  & Val.$\uparrow$  \\
\hline
\multirow{9}{*}{\rotatebox[origin=c]{90}{\textbf{LCD}}}
& LR     & $0.98 {\pm 0.01}$ & $0.65 {\pm 0.01}$ & $0.14 {\pm 0.01}$ & $0.85 {\pm 0.03}$ & $0.85 {\pm 0.02}$ & $0.23 {\pm 0.03}$ & $0.99 {\pm 0.01}$ \\
& Wac.   & $0.38 {\pm 0.03}$ & $0.76 {\pm 0.04}$ & $0.12 {\pm 0.01}$ & $0.69 {\pm 0.03}$ & $1.35 {\pm 0.05}$ & $0.26 {\pm 0.03}$ & $0.57 {\pm 0.03}$\\
& CCH.   & $0.56 {\pm 0.03}$ & $0.23 {\pm 0.02}$ & $0.23 {\pm 0.01}$ & $0.70 {\pm 0.04}$ & $0.57 {\pm 0.03}$ & $0.11 {\pm 0.01}$ & $0.99 {\pm 0.01}$\\
& REV.   & $0.58 {\pm 0.03}$ & $0.18 {\pm 0.01}$ & $0.20 {\pm 0.02}$ & $0.70 {\pm 0.01}$ & $0.89 {\pm 0.03}$ & $0.13 {\pm 0.02}$ & $0.99 {\pm 0.01}$\\
& CRU.   & $0.55 {\pm 0.03}$ & $0.02 {\pm 0.01}$ & $0.24 {\pm 0.02}$ & $0.79 {\pm 0.02}$ & $0.57 {\pm 0.02}$ & $0.13 {\pm 0.01}$ & $0.99 {\pm 0.01}$\\
& CLU.   & $0.70 {\pm 0.03}$ & $0.26 {\pm 0.01}$ & $0.31 {\pm 0.02}$ & $0.68 {\pm 0.03}$ & $0.72 {\pm 0.03}$ & $0.21 {\pm 0.02}$ & $0.81 {\pm 0.02}$\\
& FACE   & $0.72 {\pm 0.01}$ & $0.54 {\pm 0.01}$ & $0.11 {\pm 0.01}$ & $0.70 {\pm 0.02}$ & $0.91 {\pm 0.02}$ & $0.21 {\pm 0.02}$ & $0.82 {\pm 0.02}$\\
& SAE    & $0.39 {\pm 0.03}$ & $0.33 {\pm 0.02}$ & $0.09 {\pm 0.01}$ & $0.71 {\pm 0.03}$ & $0.87 {\pm 0.02}$ & $0.17 {\pm 0.02}$ & $0.92 {\pm 0.02}$\\
& CounterNet & $0.37 {\pm 0.02}$ & $0.46 {\pm 0.03}$ & $0.25 {\pm 0.01}$ & $0.65 {\pm 0.04}$ & $0.99 {\pm 0.03}$ & $0.69 {\pm 0.01}$ & $0.99 {\pm 0.01}$ \\
& GPAE-ind & $0.49 {\pm 0.03}$& $0.68 {\pm 0.04}$ & $0.10 {\pm 0.03}$ & $0.69 {\pm 0.05}$ & $1.10 {\pm 0.06}$ & $0.21 {\pm 0.03}$ & $0.96 {\pm 0.03}$\\
& GPAE   & $0.36 {\pm 0.03}$ & $0.41 {\pm 0.02}$ & $0.09 {\pm 0.01}$ & $0.72 {\pm 0.02}$ & $0.85 {\pm 0.03}$ & $0.19 {\pm 0.01}$ & $0.99 {\pm 0.01}$\\
\hline
\hline
\multirow{9}{*}{\rotatebox[origin=c]{90}{\textbf{GMC}}}
& LR     & $0.31 {\pm 0.01}$ & $0.22 {\pm 0.01}$ & $0.11 {\pm 0.02}$ & $0.55 {\pm 0.02}$ & $1.08 {\pm 0.01}$ & $0.22 {\pm 0.01}$ & $0.99 {\pm 0.01}$\\
& Wac.   & $0.04 {\pm 0.01}$ &$0.24 {\pm 0.02}$ & $0.10 {\pm 0.02}$ & $0.53 {\pm 0.02}$ & $1.04 {\pm 0.02}$ & $0.07 {\pm 0.01}$ & $0.71 {\pm 0.02}$\\
& CCH.   & $0.21 {\pm 0.02}$ &$0.21 {\pm 0.1}$ & $0.12 {\pm 0.01}$ & $0.50 {\pm 0.03}$ & $1.14 {\pm 0.05}$ & $0.15 {\pm 0.01}$ & $0.77 {\pm 0.03}$\\
& REV.   & $0.23 {\pm 0.02}$ &$0.21 {\pm 0.01}$ & $0.14 {\pm 0.01}$ & $0.52 {\pm 0.02}$ & $1.18 {\pm 0.03}$ & $0.07 {\pm 0.01}$ & $0.80 {\pm 0.03}$\\
& CRU.   & $0.17 {\pm 0.01}$ &$0.14 {\pm 0.01}$ & $0.11 {\pm 0.02}$ & $0.50 {\pm 0.01}$ & $1.19 {\pm 0.04}$ & $0.06 {\pm 0.01}$ & $0.85 {\pm 0.02}$\\
& CLU.   & $0.17 {\pm 0.01}$ &$0.18 {\pm 0.01}$ & $0.07 {\pm 0.01}$ & $0.52 {\pm 0.02}$ & $1.14 {\pm 0.02}$ & $0.07 {\pm 0.02}$ & $0.81 {\pm 0.02}$ \\
& FACE   & $0.21 {\pm 0.02}$ &$0.17 {\pm 0.01}$ & $0.05 {\pm 0.01}$ & $0.52 {\pm 0.02}$ & $1.18 {\pm 0.03}$ & $0.08 {\pm 0.01}$ & $0.86 {\pm 0.02}$\\
& SAE    & $0.07 {\pm 0.02}$ &$0.10 {\pm 0.01}$ & $0.08 {\pm 0.02}$ & $0.52 {\pm 0.01}$ & $1.03 {\pm 0.02}$ & $0.07 {\pm 0.01}$ & $0.87 {\pm 0.01}$\\
& CounterNet & $0.20 {\pm 0.02}$ & $0.17 {\pm 0.02}$ & $0.10 {\pm 0.01}$ & $0.51 {\pm 0.01}$& $1.02 {\pm 0.03}$ & $0.11 {\pm 0.01}$ & $0.97 {\pm 0.01}$ \\
& GPAE-ind& $0.07 {\pm 0.01}$ & $0.25 {\pm 0.01}$ & $0.09 {\pm 0.01}$ & $0.52 {\pm 0.02}$ & $1.10 {\pm 0.02}$ & $0.13 {\pm 0.02}$ & $0.97 {\pm 0.02}$\\
& GPAE   & $0.03 {\pm 0.01}$ &$0.21 {\pm 0.02}$ & $0.05 {\pm 0.01}$ & $0.53 {\pm 0.02}$ & $1.05 {\pm 0.03}$ & $0.06 {\pm 0.01}$ & $0.99 {\pm 0.01}$\\ 
\hline
\hline
\multirow{9}{*}{\rotatebox[origin=c]{90}{\textbf{Adult}}}
& LR     & $1.25 {\pm 0.01}$ &$1.10 {\pm 0.01}$ & $0.11 {\pm 0.01}$ & $0.71 {\pm 0.02}$ & $1.15 {\pm 0.02}$ & $0.05 {\pm 0.01}$ & $0.99 {\pm 0.01}$\\
& Wac.   & $0.28 {\pm 0.02}$ &$1.15 {\pm 0.01}$ & $0.09 {\pm 0.01}$ & $0.52 {\pm 0.02}$ & $1.29 {\pm 0.02}$ & $0.03 {\pm 0.01}$ & $0.76 {\pm 0.02}$\\
& CCH.   & $0.80 {\pm 0.03}$ &$0.17 {\pm 0.02}$ & $0.22 {\pm 0.02}$ & $0.53 {\pm 0.03}$ & $2.43 {\pm 0.04}$ & $0.03 {\pm 0.01}$ & $0.72 {\pm 0.02}$\\
& REV.   & $0.98 {\pm 0.02}$ &$0.44 {\pm 0.01}$ & $0.10 {\pm 0.02}$ & $0.55 {\pm 0.03}$ & $1.08 {\pm 0.03}$ & $0.03 {\pm 0.01}$ & $0.81 {\pm 0.01}$\\
& CRU.   & $0.81 {\pm 0.01}$ &$0.19 {\pm 0.01}$ & $0.11 {\pm 0.01}$ & $0.50 {\pm 0.01}$ & $2.49 {\pm 0.02}$ & $0.03 {\pm 0.01}$ & $0.77 {\pm 0.01}$\\
& CLU.   & $0.81 {\pm 0.03}$ &$0.13 {\pm 0.02}$ & $0.04 {\pm 0.01}$ & $0.53 {\pm 0.01}$ & $1.35 {\pm 0.04}$ & $0.03 {\pm 0.01}$ & $0.81 {\pm 0.03}$\\
& FACE   & $0.89 {\pm 0.03}$ &$0.78 {\pm 0.03}$ & $0.07 {\pm 0.01}$ & $0.56 {\pm 0.02}$ & $0.98 {\pm 0.04}$ & $0.07 {\pm 0.02}$ & $0.87 {\pm 0.03}$\\
& SAE    & $0.31 {\pm 0.02}$ &$0.89 {\pm 0.02}$ & $0.08 {\pm 0.01}$ & $0.55 {\pm 0.02}$ & $1.11 {\pm 0.03}$ & $0.05 {\pm 0.01}$ & $0.89 {\pm 0.02}$\\
& CounterNet & $0.86 {\pm 0.02}$ & $0.69 {\pm 0.04}$ & $0.07 {\pm 0.02}$ & $0.59 {\pm 0.01}$ & $0.99 {\pm 0.03}$ & $0.06 {\pm 0.02}$ & $0.94 {\pm 0.02}$\\
& GPAE-ind& $0.33 {\pm 0.02}$&$0.98 {\pm 0.02}$& $0.10 {\pm 0.01}$ & $0.51 {\pm 0.01}$ & $1.20 {\pm 0.04}$& $0.10 {\pm 0.02}$ & $0.97 {\pm 0.01}$\\
& GPAE   & $0.28 {\pm 0.02}$ &$1.02 {\pm 0.03}$ & $0.05 {\pm 0.01}$ & $0.54 {\pm 0.04}$ & $0.98 {\pm 0.04}$ & $0.04 {\pm 0.01}$ & $0.99 {\pm 0.01}$\\
\hline
\hline
\multirow{9}{*}{\rotatebox[origin=c]{90}{\textbf{FICO}}}
& LR     & $0.79 {\pm 0.01}$ &$0.78 {\pm 0.01}$ & $0.13 {\pm 0.01}$ & $0.57 {\pm 0.01}$ & $1.10 {\pm 0.02}$ & $0.08 {\pm 0.02}$ & $0.99 {\pm 0.01}$\\
& Wac.   & $0.09 {\pm 0.02}$ &$0.79 {\pm 0.02}$ & $0.11 {\pm 0.02}$ & $0.54 {\pm 0.03}$ & $1.11 {\pm 0.02}$ & $0.09 {\pm 0.01}$ & $0.81 {\pm 0.02}$\\
& CCH.   & $0.66 {\pm 0.02}$ &$0.70 {\pm 0.01}$ & $0.18 {\pm 0.02}$ & $0.51 {\pm 0.01}$ & $1.14 {\pm 0.02}$ & $0.21 {\pm 0.02}$ & $0.99 {\pm 0.01}$\\
& REV.   & $0.63 {\pm 0.01}$ &$0.68 {\pm 0.02}$ & $0.10 {\pm 0.02}$ & $0.51 {\pm 0.02}$ & $1.13 {\pm 0.03}$& $0.08 {\pm 0.02}$ & $0.87 {\pm 0.03}$\\
& CRU.   & $0.58 {\pm 0.01}$ &$0.08 {\pm 0.01}$ & $0.08 {\pm 0.02}$ & $0.54 {\pm 0.03}$ & $0.94 {\pm 0.02}$ & $0.06 {\pm 0.01}$ & $0.99 {\pm 0.01}$\\
& CLU.   & $0.59 {\pm 0.02}$ &$0.34 {\pm 0.01}$ & $0.09 {\pm 0.01}$ & $0.52 {\pm 0.01}$ & $1.18 {\pm 0.02}$ & $0.03 {\pm 0.01}$ & $0.88 {\pm 0.02}$\\
& FACE   & $0.69 {\pm 0.03}$ &$0.47 {\pm 0.02}$ & $0.07 {\pm 0.02}$ & $0.52 {\pm 0.03}$ & $1.20 {\pm 0.02}$ & $0.03 {\pm 0.01}$ & $0.87 {\pm 0.02}$\\
& SAE    & $0.11 {\pm 0.01}$ &$0.78 {\pm 0.02}$ & $0.06 {\pm 0.01}$ & $0.51 {\pm 0.01}$ & $1.12 {\pm 0.02}$ & $0.05 {\pm 0.01}$ & $0.81 {\pm 0.02}$\\
& CounterNet& $0.21 {\pm 0.02}$ &$0.71 {\pm 0.02}$ & $0.07 {\pm 0.01}$ & $0.52 {\pm 0.02}$ & $1.19 {\pm 0.03}$ & $0.07 {\pm 0.01}$ & $0.96 {\pm 0.02}$\\
& GPAE-ind& $0.13 {\pm 0.02}$&$0.78 {\pm 0.03}$& $0.10 {\pm 0.02}$ & $0.51 {\pm 0.01}$ & $1.16 {\pm 0.02}$ & $0.09 {\pm 0.01}$ & $0.97 {\pm 0.02}$\\
& GPAE   & $0.11 {\pm 0.01 }$ &$0.79 {\pm 0.02}$ & $0.05 {\pm 0.01}$ & $0.52 {\pm 0.01}$ & $1.09 {\pm 0.02}$ & $0.04 {\pm 0.01}$ & $0.99 {\pm 0.01}$\\
\hline
\hline
\multirow{9}{*}{\rotatebox[origin=c]{90}{\textbf{LAW}}}
& LR    & $1.01 {\pm 0.01}$ &$1.20 {\pm 0.01}$ & $0.17 {\pm 0.01}$& $0.52 {\pm 0.02}$ & $1.88 {\pm 0.02}$ & $0.12 {\pm 0.01}$ & $0.98 {\pm 0.02}$ \\
& Wac.  & $0.17 {\pm 0.02}$ &$1.22 {\pm 0.04}$ & $0.13 {\pm 0.02}$ & $0.52 {\pm 0.01}$ & $1.73 {\pm 0.03}$ & $0.12 {\pm 0.01}$ & $0.80 {\pm 0.02}$\\
& CCH.  & $0.99 {\pm 0.01}$ &$0.20 {\pm 0.02}$ & $0.07 {\pm 0.01}$ & $0.59 {\pm 0.01}$ & $0.95 {\pm 0.03}$ & $0.09 {\pm 0.01}$ & $0.99 {\pm 0.01}$\\
& REV.  & $0.71 {\pm 0.01}$ &$0.91 {\pm 0.03}$ & $0.06 {\pm 0.02}$ & $0.55 {\pm 0.01}$ & $1.56 {\pm 0.03}$ & $0.11 {\pm 0.01}$ & $0.81 {\pm 0.01}$\\
& CRU.  & $0.81 {\pm 0.01}$ &$0.06 {\pm 0.01}$ & $0.08 {\pm 0.01}$ & $0.63 {\pm 0.03}$ & $1.17 {\pm 0.01}$ & $0.12 {\pm 0.02}$ & $0.99 {\pm 0.01}$\\
& CLU.  & $0.79 {\pm 0.01}$ &$0.37 {\pm 0.02}$ & $0.07 {\pm 0.01}$ & $0.60 {\pm 0.03}$ & $1.21 {\pm 0.03}$ & $0.06 {\pm 0.02}$ & $0.99 {\pm 0.01}$\\
& FACE  & $0.81 {\pm 0.03}$ &$0.83 {\pm 0.02}$ & $0.03 {\pm 0.01}$ & $0.53 {\pm 0.01}$ & $1.31 {\pm 0.05}$ & $0.11 {\pm 0.01}$ & $0.81 {\pm 0.02}$\\
& SAE   & $0.39 {\pm 0.02}$&$1.17 {\pm 0.01}$ & $0.05 {\pm 0.01}$ & $0.51 {\pm 0.02}$ & $1.19 {\pm 0.01}$ & $0.10 {\pm 0.01}$ & $0.98 {\pm 0.02}$\\
& CounterNet & $0.79 {\pm 0.03}$ & $0.91 {\pm 0.03}$ & $0.07 {\pm 0.02}$ & $0.51 {\pm 0.02}$ & $0.97 {\pm 0.04}$ & $0.08 {\pm 0.01}$ & $0.99 {\pm 0.01}$\\
& GPAE-ind&$0.41 {\pm 0.01}$& $1.18 {\pm 0.04}$ & $0.10 {\pm 0.02}$ & $0.53 {\pm 0.02}$ & $1.09 {\pm 0.04}$ & $0.15 {\pm 0.03}$ & $0.97 {\pm 0.02}$\\
& GPAE  & $0.37 {\pm 0.01}$ &$1.18 {\pm 0.02}$ & $0.05 {\pm 0.01}$ & $0.52 {\pm 0.01}$ & $1.01 {\pm 0.02}$ & $0.09 {\pm 0.01}$ & $0.99 {\pm 0.01}$\\
\hline
\hline
\hline
\caption{Counterfactual evaluation with no mask. We evaluate our model and compare it with other benchmarks in L2 distance(L2), Diversity(Div.), Instability(Instb.), Discriminative Power(Dispo.), IM1, IM2 and validity. The arrow beside each metric indicates that the higher/lower it is, the better it will be. }
\label{tab:cf_eval}
\end{longtable}
\endgroup

More importantly, the counterfactual evaluations across various datasets (LCD, GMC, Adult, FICO, and LAW) provide insights into the strengths and weaknesses of different models in generating plausible and diverse counterfactuals. Among the models, the \textbf{GPAE} model consistently demonstrates a balanced performance across all datasets, highlighting its overall effectiveness. We note that no model can achieve the best scores across all the metrics. For example, a lower L2 distance might lead to a high IM1 and IM2 score. The results are shown in Table \ref{tab:cf_eval}.  Our analysis of each data set is as following. 

\paragraph{\textbf{LCD}}
The \textbf{GPAE} model shows strong performance on L2 distance and moderate Diversity. Other models like \textbf{CRU} excel in IM1 and IM2 but suffer from poor Diversity and low Instability, implying that the model tends to generate very similar counterfactuals even for very different query samples. \textbf{GPAE} strikes a balance between generating meaningful and robust counterfactuals while maintaining reasonable diversity. 

We notice that \textbf{CounterNet} achieves slightly better diversity but fails to record better IM1 and IM2 scores. In addition, the uncertainty of the decoder might have led to a higher instability score.  

\paragraph{\textbf{GMC}}
In the GMC dataset, our model also produces a solid performance on L2 distance and Validity, with Diversity and Instability values that are on par with other models such as \textbf{LR} and \textbf{Wac.}. \textbf{GPAE} achieves fairly low IM1 and IM2 score, implying that the generated counterfactuals likely stay around the manifold of existing data samples.

\paragraph{\textbf{Adult}}
The \textbf{GPAE} model’s performance remains consistent in the Adult dataset, where it balances L2 distance and IM1 and IM2 while achieving one of the highest Validity score. The \textbf{FACE} model are also achieving competitive performance on IM2 and Instability. However, it could not turn all the samples into target-class and the L2 distance is among the highest ones. 

\paragraph{\textbf{FICO}}
In the FICO dataset, \textbf{GPAE} again shows the balanced strength. While models like \textbf{CRU} and \textbf{FACE} produce slightly lower IM1 and IM2, they fall short in other critical metrics like L2 distance, with \textbf{CRU} recording a very high L2 distance and low Diversity. This might imply the counterfactual samples tend to be indifferent. 

The benefit of \textbf{GPAE} is especially evident in this dataset, as it produces counterfactuals that are both reliable and diverse, avoiding the extreme values seen in models like \textbf{CRU}. 

\paragraph{\textbf{LAW}}
In the LAW dataset, \textbf{GPAE} continues to deliver well-rounded results, while some models like \textbf{CRU} and \textbf{CLU} achieve worse L2 distance. They again score a fairly low Diversity. 

\paragraph{\textbf{Discussion}}
The overall results show that the \textbf{GPAE} model consistently delivers a balanced performance across different datasets, demonstrating strong Instability and Diversity scores. Importantly, GPAE achieves some of the lowest IM1 and IM2 scores across most datasets, indicating that its generated counterfactuals remain closer to the manifold of the target data, making them more realistic and relevant. However, we argue that IM1 and IM2 are limited in evaluating these models because a lower IM1 and IM2 might just indicate the sample stays around the manifold. The preferable Interpretability scores can also be achieved when the model tends to generate indistinguishable samples. For example, the \textbf{REV.} The model in the Adult data set has lower IM1 and IM2, but the Diversity score suggests that an invariant counterfactual might be proposed regardless of the query sample. 

The benefit of deploying a decoder during the training can be understood by comparing \textbf{GPAE} and \textbf{Wac.}. \textbf{Wac.} trains a classifier (which is the same as our encoder) and searches for a counterfactual by back-propagating the gradients. While our framework also relies on back-propagating gradients, the counterfactual samples from our model benefit from regularizing the latent space with the decoder during training and the density estimator during searching. 

In addition, models like \textbf{CRU} and \textbf{FACE} often exhibit high L2 distance or Diversity but suffer from elevated IM1 and IM2 scores, suggesting that their counterfactuals may stay further from the original data distribution. This makes GPAE a robust choice for generating meaningful and valid counterfactuals, as it balances diversity and stability while keeping the samples well-grounded within the data manifold.

\subsection{Counterfactual evaluations with mask}
\begingroup
\setlength{\tabcolsep}{4pt} 
\scriptsize
\begin{longtable}{c c|c|c|c|c|c|c|c }
\hline\hline
\multicolumn{9}{c}{\textbf{Counterfactual Evaluations with Immutable Features}}\\
\hline
\multicolumn{2}{c|}{\textbf{Model}} & L2$\downarrow$& Div.$\uparrow$ & Instb.$\downarrow$ & Dispo.$\uparrow$ & IM1$\downarrow $ & IM2$\downarrow $  & Val.$\uparrow$ \\
\hline
\multirow{8}{*}{\rotatebox[origin=c]{90}{\textbf{LCD}}} 
& Wac.   & $0.34 {\pm 0.02}$ & $0.75 {\pm 0.03}$ & $0.12 {\pm 0.01}$ & $0.73 {\pm 0.04}$ & $1.04 {\pm 0.05}$ & $0.27 {\pm 0.02}$ & $0.74 {\pm 0.02}$\\
& CCH.   & $0.51 {\pm 0.03}$ & $0.36 {\pm 0.02}$ & $0.28 {\pm 0.03}$ & $0.70 {\pm 0.02}$ & $0.64 {\pm 0.02}$ & $0.16 {\pm 0.01}$ & $0.98 {\pm 0.01}$\\
& REV.   & $0.53 {\pm 0.03}$ & $0.33 {\pm 0.02}$ & $0.19 {\pm 0.01}$ & $0.71 {\pm 0.04}$ & $0.82 {\pm 0.03}$ & $0.19 {\pm 0.02}$ & $0.98 {\pm 0.01}$\\
& CRU.   & $0.65 {\pm 0.02}$ & $0.28 {\pm 0.01}$ & $0.15 {\pm 0.01}$ & $0.75 {\pm 0.03}$ & $0.43 {\pm 0.02}$ & $0.20 {\pm 0.02}$ & $0.99 {\pm 0.01}$\\
& CLU.   & $0.49 {\pm 0.02}$ & $0.38 {\pm 0.03}$ & $0.24 {\pm 0.01}$ & $0.69 {\pm 0.03}$ & $0.92 {\pm 0.02}$ & $0.15 {\pm 0.01}$ & $0.81 {\pm 0.03}$\\
& FACE   & $0.69 {\pm 0.01}$ & $0.55 {\pm 0.01}$ & $0.17 {\pm 0.01}$ & $0.70 {\pm 0.01}$ & $0.79 {\pm 0.01}$ & $0.20 {\pm 0.01}$ & $0.87 {\pm 0.01}$\\
& CounterNet & $0.35 {\pm 0.02}$ & $0.45 {\pm 0.03}$ & $0.25 {\pm 0.02}$ & $0.71 {\pm 0.04}$ & $1.09 {\pm 0.05}$ & $0.88 {\pm 0.04}$  & $0.99 {\pm 0.01}$ \\
& GPAE   & $0.33 {\pm 0.01}$ & $0.53 {\pm 0.02}$ & $0.07 {\pm 0.01}$ & $0.74 {\pm 0.03}$ & $0.74 {\pm 0.02}$ & $0.22 {\pm 0.01}$ & $0.99 {\pm 0.01}$\\
\hline
\hline
\multirow{8}{*}{\rotatebox[origin=c]{90}{\textbf{GMC}}} 
& Wac.   & $0.04 {\pm 0.02}4$ &$0.23 {\pm 0.03}$ & $0.10 {\pm 0.02}$ & $0.52 {\pm 0.03}$ & $1.13 {\pm 0.05}$ & $0.13 {\pm 0.02}$ & $0.60 {\pm 0.03}$\\
& CCH.   & $0.17 {\pm 0.02}$ & $0.21 {\pm 0.01}$ & $0.11 {\pm 0.01}$ & $0.51 {\pm 0.01}$ & $1.19 {\pm 0.02}$ & $0.15 {\pm 0.01}$ & $0.61 {\pm 0.01}$\\
& REV.   & $0.17 {\pm 0.01}$ & $0.21 {\pm 0.01}$ & $0.12 {\pm 0.01}$ & $0.51 {\pm 0.02}$ & $1.10 {\pm 0.03}$ & $0.17 {\pm 0.01}$ & $0.70 {\pm 0.02}$ \\
& CRU.   & $0.12 {\pm 0.01}$ & $0.18 {\pm 0.01}$ & $0.09 {\pm 0.01}$ & $0.53 {\pm 0.01}$ & $1.12 {\pm 0.03}$ & $0.24 {\pm 0.02}$ & $0.73 {\pm 0.03}$\\
& CLU.   & $0.11 {\pm 0.01}$ & $0.20 {\pm 0.01}$ & $0.08 {\pm 0.01}$ & $0.52 {\pm 0.02}$ & $1.32 {\pm 0.03}$ & $0.13 {\pm 0.01}$ & $0.76 {\pm 0.03}$\\
& FACE   & $0.09 {\pm 0.02}$ & $0.16 {\pm 0.02}$ & $0.07 {\pm 0.01}$ & $0.52 {\pm 0.02}$ & $1.03 {\pm 0.03}$ & $0.13 {\pm 0.01}$ & $0.75 {\pm 0.02}$\\
& CounterNet & $0.20 {\pm 0.03}$ & $0.17 {\pm 0.02}$ & $0.10 {\pm 0.01}$ & $0.51 {\pm 0.02}$ & $1.09 {\pm 0.03}$ & $0.16 {\pm 0.01}$ & $0.90 {\pm 0.03}$\\
& GPAE   & $0.04 {\pm 0.01}$ & $0.23 {\pm 0.02}$ & $0.07 {\pm 0.01}$ & $0.52 {\pm 0.02}$ & $0.98 {\pm 0.03}$ & $0.16 {\pm 0.01}$ & $0.98 {\pm 0.02}$\\
\hline
\hline
\multirow{8}{*}{\rotatebox[origin=c]{90}{\textbf{Adult}}} 
& Wac.   & $0.28 {\pm 0.03}$ & $1.01 {\pm 0.04}$ & $0.15 {\pm 0.03}$ & $0.57 {\pm 0.04}$ & $1.00 {\pm 0.05}$ & $0.07 {\pm 0.02}$ & $0.48 {\pm 0.04}$ \\
& CCH.   & $0.62 {\pm 0.02}$ & $0.72 {\pm 0.02}$ & $0.17 {\pm 0.01}$ & $0.53 {\pm 0.03}$ & $1.11 {\pm 0.04}$ & $0.11 {\pm 0.01}$ & $0.35 {\pm 0.02}$\\
& REV.   & $0.78 {\pm 0.02}$ & $0.78 {\pm 0.02}$ & $0.09 {\pm 0.01}$ & $0.57 {\pm 0.01}$ & $1.11 {\pm 0.03}$ & $0.07 {\pm 0.01}$ & $0.47 {\pm 0.04}$\\
& CRU.   & $0.60 {\pm 0.02}$ & $0.73 {\pm 0.02}$ & $0.11 {\pm 0.01}$ & $0.51 {\pm 0.01}$ & $1.57 {\pm 0.02}$ & $0.07 {\pm 0.01}$ & $0.42 {\pm 0.03}$ \\
& CLU.   & $0.61 {\pm 0.03}$ & $0.71 {\pm 0.04}$ & $0.07 {\pm 0.01}$ & $0.53 {\pm 0.04}$ & $1.14 {\pm 0.03}$ & $0.06 {\pm 0.01}$ & $0.45 {\pm 0.02}$\\
& FACE   & $0.85 {\pm 0.02}$ & $0.79 {\pm 0.02}$ & $0.08 {\pm 0.01}$ & $0.56 {\pm 0.01}$ & $1.02 {\pm 0.02}$ & $0.06 {\pm 0.01}$ & $0.44 {\pm 0.05}$\\
& CounterNet & $0.86 {\pm 0.02}$ & $0.69 {\pm 0.02}$ & $0.07 {\pm 0.01}$ & $0.58 {\pm 0.03}$ & $1.03 {\pm 0.03}$ & $0.10 {\pm 0.02}$ & $0.84 {\pm 0.03}$\\
& GPAE   & $0.19 {\pm 0.01}$ & $1.08 {\pm 0.03}$ & $0.06 {\pm 0.01}$ & $0.59 {\pm 0.03}$ & $1.00 {\pm 0.03}$ & $0.06 {\pm 0.02}$ & $0.83 {\pm 0.03}$\\
\hline
\hline
\multirow{8}{*}{\rotatebox[origin=c]{90}{\textbf{LAW}}}
& Wac.  & $0.26 {\pm 0.03}$ & $1.12 {\pm 0.03}$ & $0.13 {\pm 0.03}$ & $0.57 {\pm 0.02}$ & $1.54 {\pm 0.04}$ & $0.14 {\pm 0.03}$ & $0.39 {\pm 0.03}$\\
& CCH.  & $0.89 {\pm 0.03}$ & $0.75 {\pm 0.02}$ & $0.05 {\pm 0.01}$ & $0.62 {\pm 0.02}$ & $1.43 {\pm 0.03}$ & $0.13 {\pm 0.01}$ & $0.99 {\pm 0.01}$\\
& REV.  & $0.79 {\pm 0.01}$ & $1.02 {\pm 0.02}$ & $0.09 {\pm 0.02}$ & $0.57 {\pm 0.01}$ & $1.37 {\pm 0.03}$ & $0.11 {\pm 0.02}$ & $0.60 {\pm 0.02}$\\
& CRU.  & $0.94 {\pm 0.03}$ & $0.53 {\pm 0.01}$ & $0.08 {\pm 0.01}$ & $0.69 {\pm 0.02}$ & $0.28 {\pm 0.01}$ & $0.13 {\pm 0.01}$ & $0.99 {\pm 0.01}$\\
& CLU.  & $0.80 {\pm 0.02}$ & $0.68 {\pm 0.03}$ & $0.07 {\pm 0.01}$ & $0.57 {\pm 0.02}$ & $0.76 {\pm 0.04}$ & $0.09 {\pm 0.02}$ & $0.99 {\pm 0.01}$\\
& FACE  & $0.91 {\pm 0.03}$ & $0.81 {\pm 0.02}$ & $0.04 {\pm 0.01}$ & $0.60 {\pm 0.03}$ & $1.63 {\pm 0.05}$ & $0.16 {\pm 0.02}$ & $0.80 {\pm 0.03}$\\
& CounterNet & $0.79 {\pm 0.03}$ & $0.91 {\pm 0.02}$ & $0.07 {\pm 0.01}$ & $0.58 {\pm 0.03}$ & $0.96 {\pm 0.03}$ & $0.09 {\pm 0.01}$ & $0.97 {\pm 0.02}$\\
& GPAE  & $0.28 {\pm 0.01}$ & $1.19 {\pm 0.03}$ & $0.06 {\pm 0.01}$ & $0.57 {\pm 0.03}$ & $0.89 {\pm 0.02}$ & $0.07 {\pm 0.01}$ & $0.97 {\pm 0.03}$\\
\hline
\hline
\caption{Counterfactual evaluation with mask. We evaluate our model with mask and compare it with other benchmarks in L2 distance(L2), Diversity(Div.), Instability(Instb.), Discriminative Power(Dispo.), IM1 and IM2. We select \textbf{Interest Rate} of \textbf{LCD}, \textbf{Age} of \textbf{GMC}, \textbf{Gender} of \textbf{LAW} and \textbf{Race,Gender} of \textbf{Adult} as the immutable features for masking. The arrow beside each metric indicates the high/lower it is, the better it will be. }
\label{tab:cf_eval_mask}
\end{longtable}
\endgroup

We also conduct the experiments to study the results of counterfactual evaluations with immutable features across four datasets—LCD, GMC, Adult, and LAW. The results are shown in Table \ref{tab:cf_eval_mask} with the description of the immutable features. 

\paragraph{\textbf{LCD}}
We select the Interest Rate as the immutable feature, as it is unfair to ask an applicant to change the interest rate to meet the requirement. This immutable feature, unlike a typical one, is a continuous variable. Our model naturally allows for selecting any future as the immutable feature by clipping the gradients of the chosen features. While \textbf{CRU.} achieves very competitive performance in Validity, IM2 and IM1, its Diversity remains low with a high L2 distance. This suggests that the generated counterfactual samples are similar to each other. In contrast, \textbf{GPAE} has achieved better Diversity with minimum L2 distance while maintaining Validity, IM1 and IM2. 

\paragraph{\textbf{GMC}}
In this data set, we select Age as the immutable feature. We observe that \textbf{Wac.} has achieved competitive performance in L2 distance, Diviersity and Dispo. However, the higher IM1 and IM2 indicate that the generated counterfactuals do not stay around the target data manifold. This is also reflected by the low Validity score. Compared with other VAE-based models such as \textbf{CCH.}, \textbf{REVISE}, \textbf{CRU} and \textbf{CLU.}, \textbf{GPAE} has the best Validity with a significant margin and a low L2 distance.

\paragraph{\textbf{Adult}}
We select Gender and Race as the immutable features in this dataset. As the number of immutable features increases, we observe that it becomes more challenging to generate the valid counterfactuals that change their label. \textbf{Wac.} failed to produce valid counterfactuals for more than half of the query samples despite its competitive performance on other metrics. The graph-based method \textbf{FACE} also performed competitively but failed to achieve a low L2 distance and a high validity rate. This might result from the fact that it is searching for counterfactuals within the existing dataset. This mechanism has two apparent challenges: 1)The performance largely depends on the size of selected samples to build the graph and 2) if more features are selected to be masked, it further reduces the size of the graph, leading to high L2 distance and low validity score. 

\paragraph{\textbf{LAW}}
In the LAW dataset, we select gender as the immutable feature. While \textbf{CRU.} and \textbf{CLU.} have achieved very competitive performance on IM1 and IM2 with a perfect validity score, the high L2 distance and low Diviersity score appear to be a similar problem that leads to indistinguishable counterfactuals. Nonetheless, \textbf{GPAE} strikes the balance over all the metrics. 

\paragraph{\textbf{Disucssion}}
Overall, given the set of immutable features, the \textbf{GPAE} model stands out across multiple datasets for its ability to consistently deliver a balanced performance in L2 distance, Diversity, and Instability. Unlike models that excel in one metric but underperform in others (e.g. \textbf{CRU.} with high IM1 and IM2 but low Diversity), \textbf{GPAE} offers a well-rounded approach, ensuring that counterfactuals are not only minimal in changes and diverse but also robust.

One of the key advantages of \textbf{GPAE} is its low IM1/IM2 scores, which indicate that the generated counterfactuals with immutable features stay closer to the data manifold. This proximity to the original target data distribution makes the counterfactuals more realistic and trustworthy. In contrast, \textbf{CounterNet} forcibly replaces the immutable features with the original values. This might lead to higher IM1 and IM2 scores, implying out-of-distribution samples. 

Furthermore, we notice that not all baseline methods are capable of generating counterfactual samples while ensuring that the immutable features are unchanged. \textbf{FACE} heavily relies on the graph that is built from the existing dataset. When a set of immutable features is selected, the searching space is shrinking and is prone to producing less diversified samples. Other VAE-based methods such as \textbf{CCH.}, \textbf{REV.}, \textbf{CRU.} and \textbf{CLU.} cannot guarantee that the immutable features are fixed due to the uncertainty from the decoder. \textbf{Wac.} searches for a counterfactual similar to ours, which ensures that the immutable features are unchanged. However, without density consideration, it has shown less preferable IM1 and IM2 scores on many datasets. In contrast, \textbf{GPAE} guarantees that immutable features stay unchanged, demonstrating its capacity to produce fair counterfactuals without introducing bias. These properties make \textbf{GPAE} a reliable and versatile option for generating counterfactuals that are not only meaningful and varied but also adhere to the principles of fairness and robustness.

\begin{figure}
    \centering
    \includegraphics[width=1\linewidth]{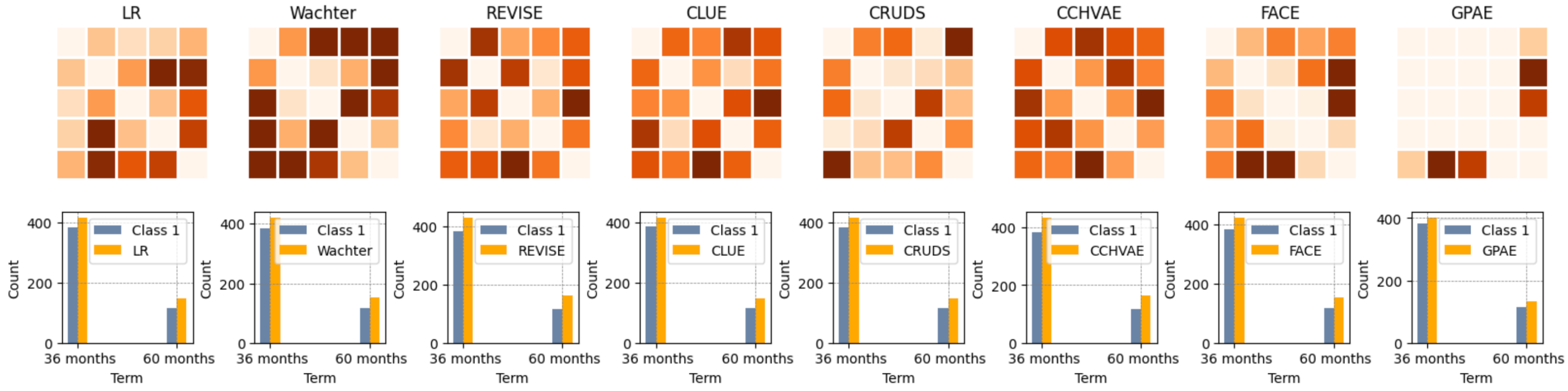}
    \caption{Qualitative comparisons between GPAE and others methods on LCD dataset. Top: the absolute difference between the correlation of counterfactual samples and that of the target class for the continuous features (debt-to-income ratio, loan amount, interest rate, annual income, FICO score). Bottom: Bar plots for the categorical variable (loan term: 36 months or 60 months).}
    \label{fig:cf_corr_compare}
\end{figure}

We also provide a qualitative comparison of the LCD dataset in Figure \ref{fig:cf_corr_compare}. LCD contains five numerical features and one categorical feature. In the non-mask setting, all features receive gradients that lead to the target class. Darker pixels show greater discrepancy between the counterfactual sample and the target class. As we can see from the first row, GPAE has fewer darker pixels in general. In the second row, we observe that GPAE is more likely to align with the target class in categorical variables. In general, the distributional agreement between the target class and the counterfactual class is much greater with GPAE than with other methods.

\subsection{Ablation study}

\begin{figure*}[t]
\centering
  \includegraphics[width=1\textwidth]{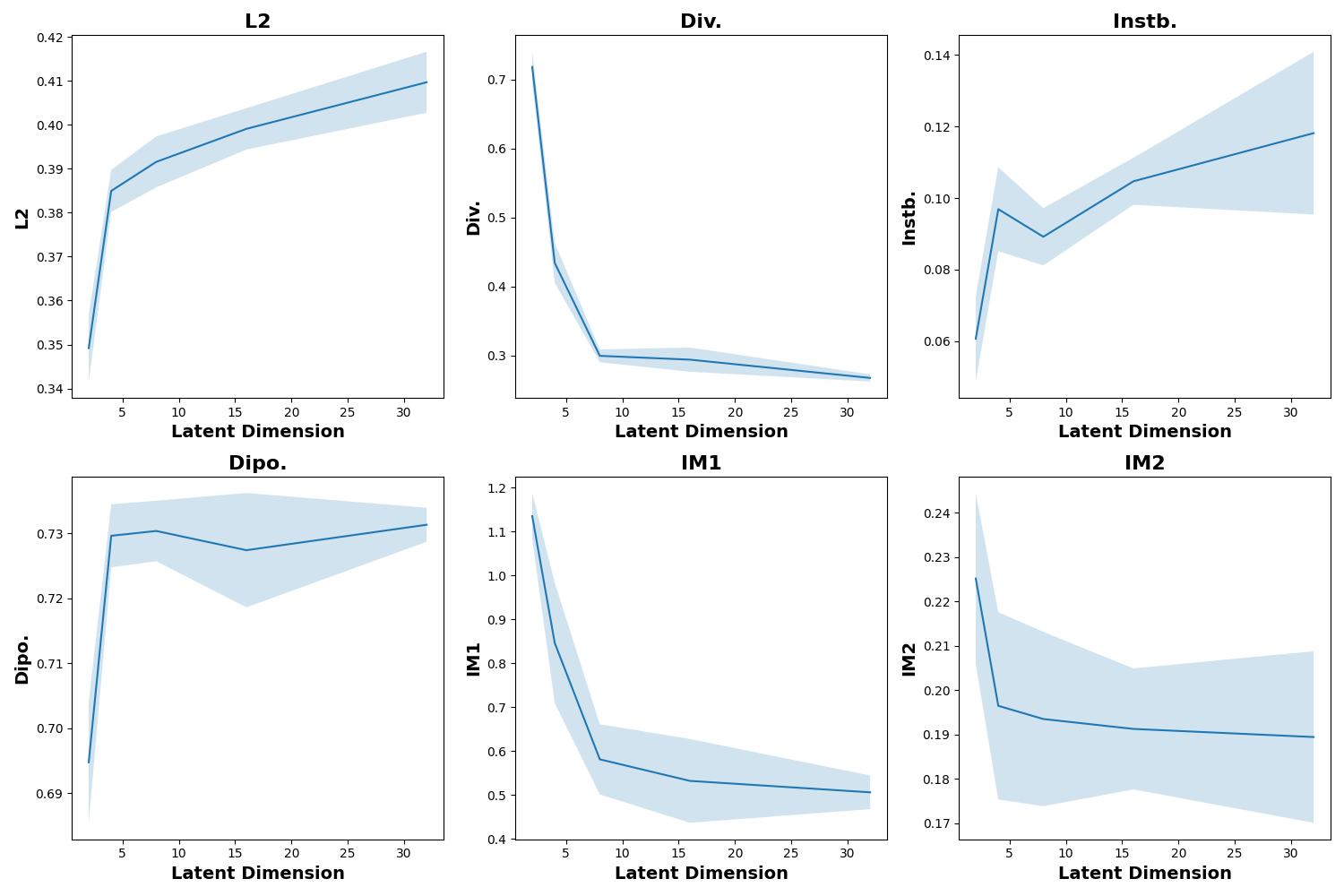}
  \caption{The relationship between different evaluation metrics and latent dimensions for the LCD dataset. The plots illustrate the trends of the evaluation metrics (L2 distance, Diversity, Instability, Discriminative power, IM1 and IM2) with the fixed RFF dimension(=1000) as the latent dimension increases. Shaded regions indicate the confidence intervals, showing the variability of each metric across different dimensions. }
  \label{fig:latent_cf}
\end{figure*}

\begin{figure*}[t]
\centering
  \includegraphics[width=1\textwidth]{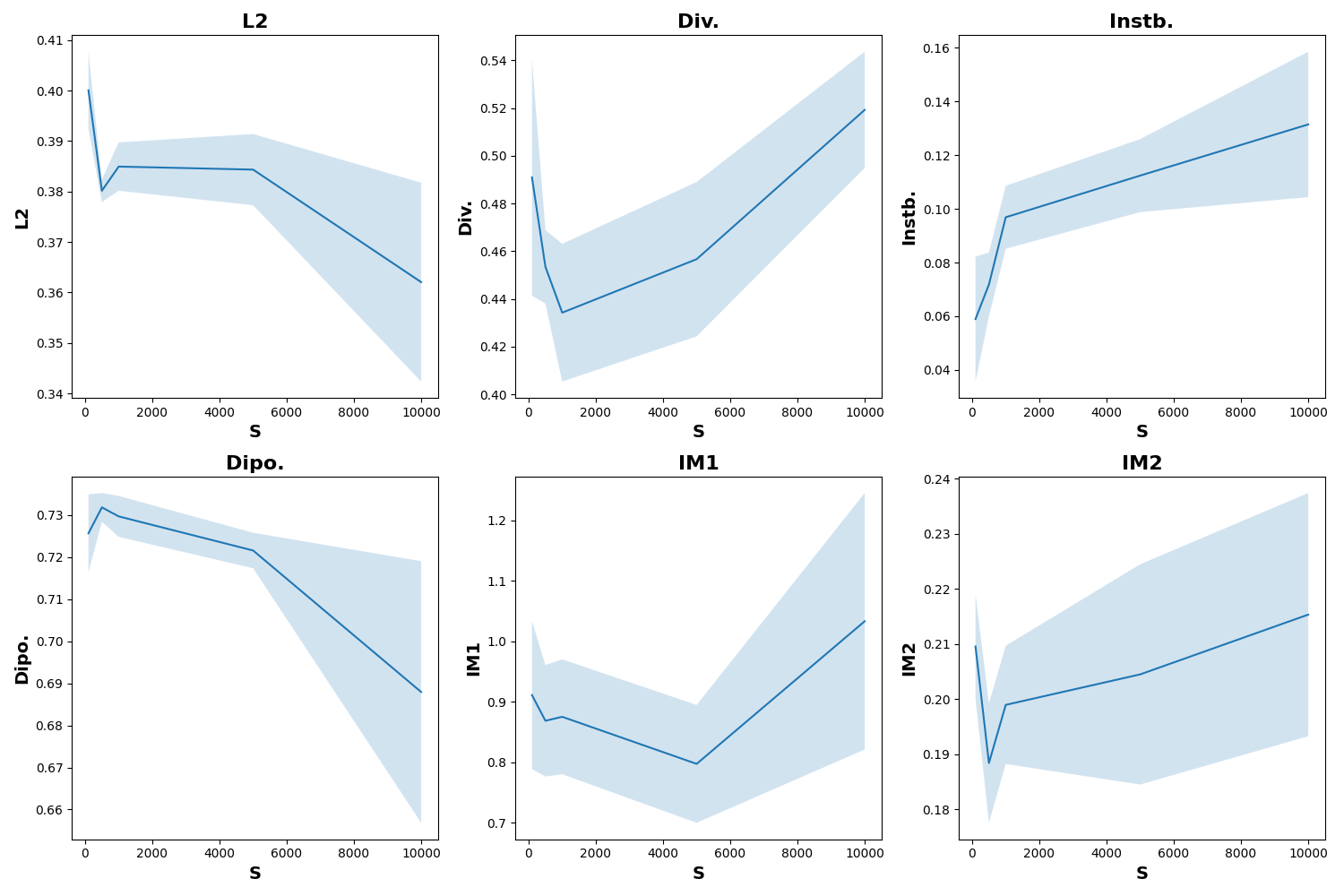}
  \caption{Evaluation of the LCD dataset using different metrics as a function of RFF dimension S with the fixed latent dimension(=4). Each plot shows the trends of L2 distance (L2), Diversity (Div.), Instability (Instb.), Discriminative power (Dipo.), IM1 and IM2 as S increases. The shaded regions represent confidence intervals, indicating variability in metric values. }
  \label{fig:S_cf}
\end{figure*}

We conduct the ablation experiments of the latent dimension and the RFF mapping dimension on counterfactual performance. In Figure \ref{fig:latent_cf}, we study the effect of the latent dimension on counterfactual quality. L2 distance rises sharply with increasing latent dimensions, indicating that larger latent spaces enable the model to generate samples that stay closer to the target distribution. This is also reflected by the IM2 and IM2, both of which decrease as the latent dimension increases, indicating the samples are more in-distribution. We notice that the diversity is decreasing, which might indicate the model tends to overfit. The Instability and Discriminative Power are improved as the latent dimension increases. Overall, although we notice that Diversity decreases, the difference in change is in an acceptable range and we observe all the other metrics improve significantly as a result. 

We also study the effect of the RFF mapping dimension on each evaluation metric. As $S$ increases, we observe notable changes in all metrics. Specifically, L2 distance shows a decreasing trend. This pattern suggests that counterfactuals bear fewer changes with increasing values of $S$ to obtain the counterfactual samples. Diversity, on the other hand, exhibits a U-shaped pattern, initially decreasing and then increasing again. Instability increases steadily as $S$ grows, indicating that counterfactuals become more sensitive to perturbations at higher values of $S$. Importantly, IM1 and IM2 values, which indicate how well counterfactuals align with the data manifold, increase for larger $S$, suggesting that high $S$ values push counterfactuals away from the original target data distribution. This trade-off indicates that moderate values of $S$ are preferred for generating diverse and plausible counterfactuals that remain close to the original data manifold.

\section{Conclusion}
In this paper, we propose a novel framework for counterfactual explanations on tabular datasets. Our framework leverage the Gaussian process to construct a supervised auto-encoder. We use RFF approximation to reduce the computational cost of Gaussian Process, leading to fewer trainable parameters and being less prone to overfitting. We further introduce a novel density estimator for searching for meaningful and in-distribution counterfactual samples. We integrate the learned density estimator in the pipeline of counterfactual searching. We also introduce the algorithm of searching for the optimal regularization rate on density estimation while searching for the counterfactuals. Our experiments on several popular tabular datasets demonstrate that our model is easier to train and is capable of generating realistic counterfactual samples.

\bibliographystyle{plainnat} 
\bibliography{ref} 

\end{document}